\def\year{2022}\relax
\documentclass[letterpaper]{article} % DO NOT CHANGE THIS
\usepackage{aaai22}  % DO NOT CHANGE THIS
\usepackage{times}  % DO NOT CHANGE THIS
\usepackage{helvet}  % DO NOT CHANGE THIS
\usepackage{courier}  % DO NOT CHANGE THIS
\usepackage[hyphens]{url}  % DO NOT CHANGE THIS
\usepackage{graphicx} % DO NOT CHANGE THIS
\usepackage{natbib}  % DO NOT CHANGE THIS AND DO NOT ADD ANY OPTIONS TO IT
\usepackage{caption} % DO NOT CHANGE THIS AND DO NOT ADD ANY OPTIONS TO IT
\DeclareCaptionStyle{ruled}{labelfont=normalfont,labelsep=colon,strut=off} % DO NOT CHANGE THIS
\usepackage{algorithm}
\usepackage{algorithmicx}
\usepackage{algpseudocode}
\usepackage{xspace}
\usepackage{amsmath,amssymb,amsthm,mathabx,amsfonts}
\usepackage{csquotes}
\usepackage{enumitem}
\usepackage{subcaption}
\usepackage[normalem]{ulem}
\usepackage{todonotes} % for comments

\usepackage{cleveref}

\makeatletter
\DeclareRobustCommand\onedot{\futurelet\@let@token\@onedot}
\def\@onedot{\ifx\@let@token.\else.\null\fi\xspace}
\def\eg{\emph{e.g}\onedot} 

\def\ie{\emph{i.e}\onedot}

\def\etc{\emph{etc}\onedot}

\makeatother

\Crefname{equation}{Eq.}{Eqs.}

\usepackage{newfloat}
\usepackage{listings}
\floatstyle{ruled}
\newfloat{listing}{tb}{lst}{}
\floatname{listing}{Listing}

\title{Energy-Based Generative Cooperative Saliency Prediction}
\author{
    Jing Zhang$^{1}$,
  Jianwen Xie$^{2}$,
  Zilong Zheng$^{3}$,
  Nick Barnes$^{1}$\\
}
\affiliations{
    $^{1}$ The Australian National University \\
  $^{2}$ Cognitive Computing Lab, 
Baidu Research \\ %\quad\\
$^{3}$ University of California, Los Angeles\\ %\quad\\
zjnwpu@gmail.com,\ \{jianwen,\  zilongzheng0318\}@ucla.edu,  nick.barnes@anu.edu.au 
}

\usepackage{bibentry}
\begin{document}
% \linenumbers

\maketitle

\begin{abstract}
Conventional saliency prediction models typically learn a deterministic mapping from an image to its saliency map, and thus fail to explain the subjective nature of human attention.  In this paper, to model the uncertainty of visual saliency, we study the saliency prediction problem from the perspective of generative models by learning a conditional probability distribution over the saliency map given an input image, and treating the saliency prediction as a sampling process from the learned distribution. Specifically, we propose a generative cooperative saliency prediction framework, where a conditional latent variable model~(LVM) and a conditional energy-based model~(EBM) are jointly trained to predict salient objects in a cooperative manner. The LVM serves as a \textit{fast but coarse} predictor to efficiently produce an initial saliency map, which is then refined by the iterative Langevin revision of the EBM that serves as a \textit{slow but fine} predictor. Such a coarse-to-fine cooperative saliency prediction strategy offers the best of both worlds. Moreover, we propose a ``cooperative learning while recovering" strategy and apply it to weakly supervised saliency prediction, where saliency annotations of training images are partially observed. Lastly, we find that the learned energy function in the EBM can serve as a refinement module that can refine the results of other pre-trained saliency prediction models. Experimental results show that our model can produce a set of diverse and plausible saliency maps of an image, and obtain state-of-the-art performance in both fully supervised and weakly supervised saliency prediction tasks.
\end{abstract}

\section{Introduction}

As a class-agnostic segmentation task, salient object detection has attracted a lot of attentions in the computer vision community for its close relationship to human visual perception.
A \textit{salient region} is a visually distinctive scene region that can be located rapidly and with little human effort. Salient object detection is commonly treated as a pixel-wise binary output of a deterministic prediction model in most recent works ~\cite{CPD_Sal,BASNet_Sal,SCRN_iccv,F3Net_aaai2020,Iter_Coop_CVPR,xu2021locate}.
Despite the success of those recent models,
the one-to-one deterministic mapping
has prevented them from modeling the uncertainty of human salient object prediction, which is considered to be subjective~\cite{itti_sal} and affected by biological factors (\eg, contrast sensitivity), contextual factors (\eg, task, experience, and interest), \etc. In this way, it is more reasonable to represent visual saliency as a conditional probability distribution over a saliency map given an input image, and formulate the saliency prediction as a stochastic sampling process from the conditional distribution.

Generative models \cite{GAN_nips,vae_bayes_kumar,xie_generative_covnet,coopnets}
% , on the other hand, 
have demonstrated their abilities to represent conditional distributions of high-dimensional data and produce multiple plausible outputs given the same input~\cite{image2image_vae, xie2019cooperative}. In this work, we fit the saliency detection task into a generative framework, where the input image is the condition, and the goal is to generate multiple saliency maps, representing the \enquote{subjective nature} of human visual saliency.  \citet{jing2020uc,ucnet_jornal} have used conditional variational auto-encoders (VAEs)~\citep{vae_bayes_kumar,structure_output}, which are latent variable models (LVMs), to implicitly represent distributions of visual saliency. However, VAEs only learn a stochastic mapping from image domain to saliency domain, and lack an intrinsic cost function to evaluate the visual saliency output and guide the saliency prediction process. As to a prediction task, a cost function of solution is more reliable than a mapping function because the former is more generalizable than the latter.

In contrast, we propose to model the conditional distribution of visual saliency \textit{explicitly} via an energy-based model~(EBM)~\cite{xie_generative_covnet,short_run_mcmc_ebm}, where the energy function defined on both image and saliency domains serves as a cost function of the saliency prediction.  Given an input image, the saliency prediction can be achieved by performing sampling from the EBM via Markov chain Monte Carlo (MCMC) \citep{mcmc_langevin} method, which is a gradient-based algorithm that searches local minima of the cost function of the EBM conditioned on the input image.

A typical high-dimensional EBM learns an energy function by MCMC-based maximum likelihood estimation (MLE), which commonly suffers from convergence difficulty and computational expensiveness of the MCMC process. Inspired by prior success of energy-based generative cooperative learning ~\cite{coopnets, xie2019cooperative}, we propose the \textit{energy-based generative cooperative saliency prediction framework}  to tackle the saliency prediction task. Specifically,
the framework consists of a conditional EBM whose energy function is parameterized by a bottom-top neural network, and a conditional LVM whose transformation function is parameterized by an encoder-decoder framework. The framework brings in an LVM as an ancestral sampler to initialize the MCMC computational process of the EBM for efficient sampling, so that the EBM can be learned efficiently. The EBM, in turn, refines the LVM's generated samples via MCMC and feeds them back to the LVM, so that the LVM can learn its mapping function from the MCMC transition. Thus, the resulting cooperative saliency prediction process first generates an initial saliency map via a direct mapping and then refines the saliency map via an iterative process. This is a coarse-to-fine generative saliency detection, and corresponds to a fast-thinking and slow-thinking system~\cite{xie2019cooperative}.

Moreover, based on the \textit{generative cooperative saliency prediction framework}, we further propose a \textit{cooperative learning while cooperative recovering} strategy for weakly supervised saliency learning,
% in which we learn our model from incomplete data, 
where each training image is associated with a partially observed annotation (\eg, scribble annotation \cite{jing2020weakly}). At each learning iteration, the strategy has two sub-tasks: cooperative recovery and cooperative learning. As to the cooperative recovery sub-task, each incomplete saliency ground truth is firstly recovered in the low-dimensional latent space of the LVM via inference, and then refined by being pushed to the local mode of the cost landscape of the EBM via MCMC. For the cooperative learning sub-task, the recovered saliency maps are treated as pseudo labels to update the parameters of the framework as in the scenario of learning from complete data.  

In experiments, we demonstrate that our framework can not only achieve state-of-the-art performances
in both fully supervised and weakly supervised saliency predictions, but also generate diverse saliency maps from one input image, indicating the success of modeling the uncertainty of saliency prediction. Furthermore, we show that the learned energy function in the EBM can serve as a cost function, which is useful 
to refine the results from other pre-trained  saliency prediction models.

Our contributions can be summarized as below: 
\begin{itemize}
    \item We study generative modeling of saliency prediction, and formulate it as a sampling process from a probabilistic model using EBM and LVM respectively, which are new angles to model and solve saliency prediction. 
    \item We propose a \textit{generative cooperative saliency prediction framework}, which jointly trains the LVM predictor and the EBM predictor in a cooperative learning scheme to offer reliable and efficient saliency prediction. 
    \item We generalize our generative framework to the weakly supervised saliency prediction scenario,~in which only incomplete annotations are provided, by proposing the \textit{cooperative learning while recovering} algorithm,
where we train the model and simultaneously recover the unlabeled areas of the incomplete saliency~maps. 
    \item We provide strong empirical results in both fully supevised and weakly supervised settings to verify the effectiveness of our  framework for saliency prediction.
\end{itemize}

\section{Related Work}
 We first briefly introduce existing fully supervised and weakly supervised saliency prediction models. We then review the family of generative cooperative models and other conditional deep generative frameworks. 

\textbf{Fully Supervised Saliency Prediction}. Existing fully supervised saliency prediction models \cite{Wang_2018_CVPR,picanet,F3Net_aaai2020,Liu2019PoolSal,BASNet_Sal,SCRN_iccv,CPD_Sal, PAGE_cvpr19,Iter_Coop_CVPR,wei2020label,xu2021locate} mainly focus on exploring image context information and generating structure-preserving predictions.\citet{SCRN_iccv,Iter_Coop_CVPR,PAGE_cvpr19,Wang_2018_CVPR,picanet,Liu2019PoolSal,CPD_Sal,xu2021locate} propose saliency prediction models by effectively integrating higher-level and lower-level features. \citet{F3Net_aaai2020,wei2020label} propose an edge-aware loss term to penalize errors along object boundaries.\citet{jing2020uc} present a stochastic RGB-D saliency detection network based on the conditional variational auto-encoder \cite{vae_bayes_kumar,vae2}. In this paper, we introduce the conditional cooperative learning framework \cite{coopnets,xie2019cooperative} to achieve probabilistic coarse-to-fine RGB saliency detection, where a coarse prediction is produced by a conditional latent variable model and then is refined by a conditional energy-based model. Our paper is the first work to use a deep energy-based generative framework for probabilistic saliency detection. 

\textbf{Weakly Supervised Saliency Prediction}. Weakly supervised saliency prediction frameworks \cite{imagesaliency,Guanbin_weaksalAAAI,DeepUSPSDR,jing2020weakly} attempt to learn predictive models from easy-to-obtain weak labels, including image-level labels \cite{imagesaliency,Guanbin_weaksalAAA}, noisy labels \cite{DeepUSPSDR,Zhang_2018_CVPR,Zhang_2017_ICCV} or scribble labels \cite{jing2020weakly}. In this paper, we also propose a \textit{cooperative learning while recovering} strategy for weakly supervised saliency prediction, in which only scribble labels are provided and our model treats them as incomplete data and recovers them during learning.

\textbf{Energy-Based Generative Cooperative Networks}. Deep energy-based generative models \cite{xie_generative_covnet}, with energy functions parameterized by modern convolutional neural networks, are capable of modeling the probability density of high-dimensional data. They have been applied to image generation \cite{xie_generative_covnet,learning_gconv_mgms,short_run_mcmc_ebm,implicit_ebm,your_classifier_ebm,ZhaoXL21,ZhengXL21}, video generation \cite{xie2019learning}, 3D volumetric shape generation \cite{xie2018learning,Xie2020GenerativeVL}, and unordered point cloud generation \cite{xie2021GPointNet}. The maximum likelihood learning of the energy-based model typically requires iterative MCMC sampling, which is computationally challenging. To relieve the computational burden of MCMC, the Generative Cooperative Networks (CoopNets) in \citet{coopnets} propose to learn a separate latent variable model (\ie~a generator) to serve as an efficient approximate sampler for training the energy-based model.
\citet{xie2020learning} propose a variant of CoopNets by replacing the generator with a variational auto-encoder (VAE) \cite{VAE1}. 
\citet{xie2019cooperative} propose a conditional CoopNets for supervised image-to-image translation. Our paper proposes a conditional CoopNets for visual saliency prediction. Further, we generalize our model to the weakly supervised learning scenario by proposing a \textit{cooperative learning while recovering} algorithm. In this way, we can learn from incomplete data for weakly supervised saliency prediction.   

\textbf{Conditional Deep Generative Models}. Our framework belongs to the family of conditional generative models, which include conditional generative adversarial networks (CGANs) \cite{conditional_gan} and conditional variational auto-encoders (CVAEs) \cite{structure_output}. Different from existing CGANs \cite{semantic_gan_nips_2016, SegGAN, SegAN_2017, Pan_2017_SalGAN,Sal_CGAN,Hung_semiseg_2018,semi_seg_gan}, which train a conditional discriminator and a conditional generator in an adversarial manner, or CVAEs \cite{prob_unet,jing2020uc,ucnet_jornal}, in which a conditional generator is trained with an approximate inference network, our model learns a conditional generator with a conditional energy-based model via MCMC teaching. Specifically, our model allows an additional refinement for the generator during prediction, which is lacking in both CGANs and CVAEs.

\section{Cooperative Saliency Prediction}
\label{our_method}
We~will~first present two types of generative modeling of saliecny prediction, \ie., the energy-based model (EBM) and the latent variable model (LVM). Then, we propose a novel generative saliency prediction framework, in which the EBM and the LVM are jointly trained in a generative cooperative manner, such that they can help each other for better saliency prediction in terms of computational efficiency and prediction accuracy. The latter aims to generate a coarse but fast prediction, and the former serves as a fine saliency predictor. The resulting model is a coarse-to-fine saliency prediction framework.

\subsection{EBM as a Slow but Fine Predictor} \label{sec:ebm}
Let $X$ be an image, and $Y$ be its saliency map. The EBM defines a conditional distribution of $Y$ given $X$ by:
\begin{equation}
    p_{\theta}(Y|X) = \frac{p_{\theta}(Y,X)}{\int p_{\theta}(Y,X)dY} = \frac{\exp[-U_{\theta}(Y,X)]}{Z(X;\theta)}, \label{eq:ebm}
\end{equation}
where the energy function $U_{\theta}(Y,X)$, parameterized by a bottom-up neural network, maps the input image-saliency pair to a scalar, and $\theta$ represent the network parameters.
% in the network. 
$Z(X;\theta)=\int \exp[-U_{\theta}(Y,X)]dY$ is the normalizing constant. When 
%NB the 
$U_{\theta}$ is learned and an image $X$ is given, the prediction of saliency $Y$ can be achieved by Langevin sampling \cite{mcmc_langevin}, which makes use of the gradient of the energy function and iterates the following step:
\begin{equation}
    Y_{\tau+1} = Y_{\tau} - \frac{\delta^2}{2} \frac{\partial}{\partial Y}  U_{\theta}(Y_{\tau},X) + \delta \Delta_{\tau},
\label{equ:ebm_Langevin}
\end{equation}
where $\tau$ indexes the Langevin time step, $\delta$ is the step size, and $\Delta_{\tau} \sim \mathcal{N}(0,I_D)$ is a Gaussian noise term. The Langevin dynamics \cite{mcmc_langevin} is initialized with Gaussian distribution and is equivalent to a stochastic gradient descent algorithm that seeks to find the minimum of the objective function defined by $U_{\theta}(Y, X)$. The noise term $\Delta_{\tau}$ is a Brownian motion that prevents gradient descent from
being trapped by local minima of $U_{\theta}(Y, X)$.
The energy function $U_{\theta}(Y,X)$ in Eq.~(\ref{eq:ebm}) can be regarded as a trainable cost function of the task of saliency prediction.
The prediction process via Langevin dynamics in Eq.~(\ref{equ:ebm_Langevin}) can be considered as finding $Y$ to minimize the cost $U_{\theta}(Y,X)$ given an input $X$. Such a framework can learn a reliable and generalizable cost function for saliency prediction. However, due to the iterative sampling process in the prediction, EBM is slower than LVM, which adopts a mapping for direct sampling.   

\subsection{LVM as a Fast but Coarse Predictor} \label{sec:lvm}
Let $h$ be a latent Gaussian noise vector. %, $X$ be an image, and $Y$ be its saliency map. 
The LVM defines a mapping function  $G_{\alpha}(X, h): [h,X] \rightarrow Y$ that maps~a~latent vector $h \sim \mathcal{N}(0,I_d)$ together with an image $X$ to a saliency map $Y$. $I_d$ is a $d$-dimensional~identity matrix. $d$ is the number of dimensionalities of $h$. Specifically, the mapping function $G$ is parameterized by a noise-injected encoder-decoder network with skip connections and $\alpha$ contain all the learning parameters in the network. The LVM is~given~by:
\begin{equation}
    Y=G_{\alpha}(X,h)+ \epsilon, \epsilon \sim \mathcal{N}(0,\sigma^2 I_D),\label{eq:lvm}
\end{equation}
where $\epsilon$ is an observation residual and $\sigma$ is a predefined standard deviation of $\epsilon$. The LVM in Eq. (\ref{eq:lvm}) defines an implicit conditional distribution of saliency $Y$ given an image $X$, \ie, $p_{\alpha}(Y|X)=\int p_{\alpha}(Y|X, h)p(h)dh$, 
% , \ie, $p_{\alpha}(Y|X)=\int p(h)p_{\alpha}(Y|X, h)dh$, 
where $p_{\alpha}(Y|X,h)=\mathcal{N}(G_{\alpha}(X,h), \sigma^2 I_D)$. The saliency prediction can be achieved by an ancestral sampling process that first samples a Gaussian white noise vector
% \NB{vector?} 
$h$ and then transforms it
% the noise 
along with an image $X$ to a saliency map $Y$. Since the ancestral sampling is a direct mapping, it is faster than the iterative Langevin dynamics in the EBM. However, without a cost function as in the EBM, the learned mapping in the LVM is hard to be generalized to a new domain.

\subsection{Cooperative Prediction with Two Predictors} \label{sec:coop_sampling}

We propose to predict image saliency by a cooperative sampling strategy. We first use the coarse saliency predictor (LVM) to generate an initial prediction $\hat{Y}$ via a non-iterative ancestral sampling, and then we use the fine saliency predictor (EBM) to refine the initial prediction via $K$-step Langevin revision to obtain a revised saliency $\tilde{Y}$. The process can be written as:
\begin{eqnarray}
\begin{aligned}
& \hat{Y}=G_{\alpha}(X,\hat{h}),\hat{h} \sim \mathcal{N}(0,I_d), \\ 
& \begin{cases}
\tilde{Y}_0=\hat{Y} &\\
\tilde{Y}_{\tau+1} = \tilde{Y}_{\tau} - \frac{\delta^2}{2} \frac{\partial }{\partial \tilde{Y}} U_{\theta}(\tilde{Y}_{\tau},X) + \delta 
\Delta_\tau
\end{cases}
\label{eq:coop_sampling}
\end{aligned}
\end{eqnarray}
We call this process the cooperative sampling-based coarse-to-fine prediction. In this way, we take both advantages of these two saliency predictors in the sense that the fine saliency predictor (\ie, Langevin sampler) is initialized by the efficient coarse saliency predictor (\ie., ancestral sampler), while the coarse saliency predictor is refined by the accurate fine saliency predictor that aims to minimize a cost function $U_\theta$. 

Since our conditional model represents a one-to-many mapping, the prediction is stochastic. To evaluate the learned model on saliency prediction tasks, we can draw multiple $\hat{h}$ from the prior $\mathcal{N}(0,I_d)$ and use their average to generate $\hat{Y}$, then a Langevin dynamics with the diffusion term being disabled (\ie, gradient descent) is performed to push $\hat{Y}$ to its nearest local minimum $\tilde{Y}$ based on the learned energy function. The resulting $\tilde{Y}$ is treated as a prediction of our model.

\subsection{Cooperative Training of Two Predictors}

\label{cooperative_learning_sec}
We use the cooperative training method \cite{coopnets,xie2019cooperative} to learn the parameters
% $(\theta, \alpha)$ 
of the two predictors.  At each iteration, we
% the algorithm 
first generate synthetic examples via the cooperative sampling strategy shown in Eq.~(\ref{eq:coop_sampling}), and then the synthetic examples are used to compute the learning gradients to update both predictors. We present the update formula of each predictor below.   

{\bf MCMC-based Maximum Likelihood Estimation (MLE) for the Fine Saliency Predictor}. Given a training dataset $\{(X_i,Y_i)\}_{i=1}^n$, we train the fine saliency predictor via MLE, which maximizes the log-likelihood of the data $L(\theta) = \frac{1}{n} \sum_{i=1}^{n} \log p_{\theta}(Y_i|X_i) $, whose learning gradient is $\Delta{\theta}=\frac{1}{n} \sum_{i=1}^{n} \{\mathbb{E}_{p_{\theta}(Y|X_i)}[\frac{\partial}{\partial \theta} U_{\theta}(Y,X_i)]-\frac{\partial}{\partial \theta}U_{\theta}(Y_i,X_i)\}$. We
% run $\tilde{n}$ parallel chains of cooperative sampling
rely on the cooperative sampling in Eq.~(\ref{eq:coop_sampling}) to sample $\tilde{Y_i} \sim p_{\theta}(Y|X_i)$ to approximate the gradient: 
\begin{equation}
    \Delta{\theta} \approx \frac{1}{n} \sum_{i=1}^{n} \frac{\partial}{\partial \theta}U_{\theta}(\tilde{Y}_i,X_i) -  \frac{1}{n} \sum_{i=1}^{n} \frac{\partial}{\partial \theta} U_{\theta}(Y_i,X_i).
\label{equ:ebm_update}
\end{equation}
We can use Adam \citep{KingmaB14} with $\Delta{\theta}$ to update $\theta$. We denote $\Delta{\theta}(\{Y_i\},\{\tilde{Y}_i\})$ as a function of $\{Y_i\}$ and $\{\tilde{Y}_i\}$.

{\bf Maximum Likelihood Training of  the Coarse Saliency Predictor by MCMC Teaching}. Even though the fine saliency predictor learns from the training data, the coarse saliency predictor learns to catch up with the fine saliency predictor by treating $\{(X,\tilde{Y})\}_{i=1}^n$ as training examples. The learning objective is to maximize the log-likelihood of the samples drawn from $p_{\theta}(Y|X)$, \ie, $L(\alpha)=\frac{1}{n} \sum_{i=1}^{n} \log p_{\alpha}(\tilde{Y}_i|X_i)$, whose gradient can be computed by
\begin{equation}
    \Delta{\alpha}=\sum_{i=1}^{n} \mathbb{E}_{h \sim p_{\alpha}(h|Y_i,X_i)}\left[\frac{\partial}{\partial \alpha}\log p_{\alpha}(Y_i,h|X_i)\right]. 
\end{equation}
This leads to an MCMC-based solution that iterates (i) an inference step: inferring latent $\tilde{h}$ by sampling from posterior distribution $\tilde{h} \sim p_{\alpha}(h|Y,X)$ via Langevin dynamics, which iterates the following:
\begin{equation}
\begin{aligned}
    \tilde{h}_{\tau+1} = \tilde{h}_{\tau} + \frac{\delta^2}{2} \frac{\partial}{\partial \tilde{h}} \log p_{\alpha}(Y, \tilde{h}_{\tau}|X) + \delta \Delta_{\tau}, \label{equ:LVM_Langevin}
\end{aligned}
\end{equation}
where $\Delta_{\tau} \sim \mathcal{N}(0,I_d)$ and $\frac{\partial}{\partial \tilde{h}} \log p_{\alpha}(Y, \tilde{h}|X)=\frac{1}{\sigma^2}(Y - G_{\alpha}(X,\tilde{h})) \frac{\partial}{\partial \tilde{h}} G_{\alpha}(X, \tilde{h}) - \tilde{h}$,
and (ii) a learning step: with $\{\tilde{h}_i, \tilde{Y}_i, X_i\}$, we update $\alpha$ via Adam  optimizer with  
\begin{equation}
    \Delta{\alpha} \approx \frac{1}{n} \sum_{i=1}^{n} \frac{1}{\sigma^2}(\tilde{Y}_i - G_{\alpha}(X_i, \tilde{h}_i)) \frac{\partial}{\partial \alpha} G_{\alpha}(X_i, \tilde{h}_i). \label{equ:LVM_update}
\end{equation}

Since $G_{\alpha}$ is parameterized by a differentiable neural network, both $\frac{\partial}{\partial h} G_{\alpha}(X, h) $ in Eq.~(\ref{equ:LVM_Langevin}) and $\frac{\partial}{\partial \alpha} G_{\alpha}(X_i, \tilde{h}_i)$ in Eq.~(\ref{equ:LVM_update}) can be efficiently computed by back-propagation.
We denote $\Delta{\alpha}(\{\tilde{h}_i\},\{\tilde{Y}_i\})$ as a function of $\{\tilde{h}_i\}$ and $\{\tilde{Y}_i\}$.
Algorithm \ref{alg1} presents a description of the cooperative learning algorithm of the fine and coarse saliency predictors. 

\begin{algorithm}[H]
\small
\caption{Training the
% algorithm for 
Cooperative Saliency Predictor}
\textbf{Input}: \\
(1) Training images $\{X_i\}_{i}^{n}$ with associated saliency maps $\{Y_i\}_{i}^{n}$;\\
(2) maximal number of learning iterations $T$. 

\textbf{Output}: 
Parameters $\theta$ and $\alpha$ 
\begin{algorithmic}[1]
\State Initialize $\theta$ and $\alpha$ with Gaussian noise
\For{$t \leftarrow  1$ to $T$}
\State Draw $\hat{h}_i \sim \mathcal{N}(0,I_d)$
\State Sample initial prediction $\hat{Y}_i = G_{\alpha}(X_i,\hat{h}_i)$.
\State Revise $\hat{Y}_i$ to obtain $\tilde{Y}_i$ by Langevin dynamics in Eq. (\ref{equ:ebm_Langevin})
\State Revise $\hat{h}_i$ to obtain $\tilde{h}_i$ by Langevin dynamics in Eq. (\ref{equ:LVM_Langevin}) 
\State Update $\theta$ with $\Delta \theta(\{Y_i\},\{\tilde{Y}_i\})$ in Eq.~(\ref{equ:ebm_update}) using Adam
\State Update $\alpha$ with $\Delta \alpha(\{\tilde{h}_i\},\{\tilde{Y}_i\})$ in Eq.~(\ref{equ:LVM_update}) using Adam
\EndFor
\end{algorithmic} \label{alg1}
% \vspace{5mm}
\end{algorithm}

\section{Weakly Supervised Saliency Prediction}
\label{weak_sec}
In Section \ref{our_method}, the framework is trained from fully-observed training data. In this section, we want to show that our generative framework can be modified to handle the  scenario in which each image $X_i$ only has a partial pixel-wise annotation $Y'_i$, \eg, scribble annotation \cite{jing2020weakly}. 
Since the saliency map for each training image is incomplete, directly applying the algorithm to the incomplete training data can lead to a failure of learning the distribution of saliency given an image. However, generative models are good at data recovery, therefore they can learn to recover the incomplete data. In our framework, we will leverage the recovery powers of both EBM and LVM to deal with the incomplete data in our cooperative learning algorithm, and this will lead to a novel weakly supervised saliency prediction framework.

To learn from incomplete data, our algorithm alternates the \textit{cooperative learning step} and the \textit{cooperative recovery step}. Both steps need a cooperation between EBM and LVM. The cooperative learning step is the same as the one used for fully observed data, except that it treats the recovered saliency maps, which are generated from the cooperative recovery step, as training data in each iteration. The following is the cooperative recovery step, which consists of two sub-steps driven by the LVM and the EBM respectively:

{\bf (i) Recovery by LVM in Latent Space}. Given an~image $X_i$ and its incomplete saliency map $Y'_i$, the recovery of the missing part of $Y'_i$ can be achieved by first inferring the latent vector $h'_i$ based on the partially observed saliency information via $h'_i \sim p_{\alpha}(h|Y'_i,X_i)$, and then generating $\hat{Y}'_i=G_{\alpha}(X_i, h'_i)$ with the inferred latent vector 
$h'_i$. Let $O_i$ be a binary mask, with the same size as $Y'$, indicating the locations of visible annotations in $Y'_i$. $O_i$ varies for different $Y'_i$ and can be extracted from $Y'_i$. The Langevin dynamics %in the latent space 
for recovery iterates the same step in Eq.~(\ref{equ:LVM_Langevin}) except that
$\frac{\partial}{\partial h} \log p_{\alpha}(Y', h_{\tau}|X)=\frac{1}{\sigma^2}( O\circ (Y - G_{\alpha}(X,h_{\tau}))) \frac{\partial}{\partial h} G_{\alpha}(X, h_{\tau}) - h_{\tau}$, where $\circ$ denotes element-wise matrix multiplication operation.

{\bf (ii) Recovery by EBM in Data Space}. With the initial recovered result $\hat{Y}'$ generated by the coarse saliency predictor $p_{\alpha}$, the fine saliency predictor $p_{\theta}$ can further refine the result by running a finite-step Langevin dynamics, which is initialized with $\hat{Y}'$, to obtain $\tilde{Y}'$. 
The underlying principle is that the initial
recovery $\hat{Y}'$ might be just around one local mode of the energy function. A few steps of Langevin dynamics (\ie, stochastic gradient descent) toward $p_{\theta}$, starting from $\hat{Y}_i'$, will push $\hat{Y}_i'$ to its nearby low energy mode, which might correspond to its complete version $Y_i$.  

\begin{algorithm}[H]
\small
\caption{Cooperative learning while recovering}
\label{alg2}
\textbf{Input}: \\
(1) Images $\{X_i\}_{i}^{n}$ with incomplete annotations $\{Y'_i\}_{i}^{n}$;\\
(2) Number of learning iterations $T$ 

\textbf{Output}:
Parameters $\theta$ and $\alpha$ 

\begin{algorithmic}[1]
\State Initialize $\theta$ and $\alpha$ with Gaussian noise
\For{$t \leftarrow  1$ to $T$}
\State Infer $\hat{h}'_i$ from the visible part of $Y'_i$ by Langevin dynamics in Eq. (\ref{equ:LVM_Langevin}) 
\State Obtain initial recovery $\hat{Y}_i = G_{\alpha}(X_i,\hat{h}'_i)$.
\State Revise $\hat{Y}'_i$ to obtain $\tilde{Y}'_i$ by Langevin dynamics in Eq. (\ref{equ:ebm_Langevin})
\State Draw $\hat{h}_i \sim \mathcal{N}(0,I_d)$
\State Sample initial prediction $\hat{Y}_i = G_{\alpha}(X_i,\hat{h}_i)$.
\State Revise $\hat{Y}_i$ to obtain $\tilde{Y}_i$ by Langevin dynamics in Eq. (\ref{equ:ebm_Langevin})
\State Revise $\hat{h}_i$ to obtain $\tilde{h}_i$ by Langevin dynamics in Eq. (\ref{equ:LVM_Langevin}) 
\State Update $\theta$ with $\Delta \theta(\{\tilde{Y}'_i\},\{\tilde{Y}_i\})$ using Adam
\State Update $\alpha$ with $\Delta \alpha(\{\tilde{h}_i\},\{\tilde{Y}_i\})$ using Adam
\EndFor
\end{algorithmic} 
% \vspace{5mm}
\end{algorithm}

{\bf Cooperative Learning and Recovering}. At each iteration $t$, we perform the above cooperative recovery of the incomplete saliency maps $\{Y'\}_{i=1}^n$ via $p_{\theta^{(t)}}$ and $p_{\alpha^{(t)}}$, while learning $p_{\theta^{(t+1)}}$ and $p_{\alpha^{(t+1)}}$ from $\{X_i, \tilde{Y}_i'^{(t)}\}_{i=1}^n$, where $\{\tilde{Y}_{i}'^{(t)}\}_{i=1}^n$ are the recovered saliency maps at iteration $t$. The parameters $\theta$ are still updated via Eq.~(\ref{equ:ebm_update}) except that we replace $Y_i$ by $\tilde{Y}'_i$. That is, at each iteration, we use the recovered $\tilde{Y}'_i$, as well as the synthesized $\tilde{Y}_i$, to compute the gradient of the log-likelihood, which is denoted by $\Delta \theta(\{\tilde{Y}'_i\},\{\tilde{Y}_i\})$.
The algorithm simultaneously performs (i) cooperative recovering of missing annotations of each training example; 
 (ii) cooperative sampling to generate annotations; (iii) cooperative learning of the two models by updating parameters with both recovered annotations and generated annotations. See Algorithm \ref{alg2} for a detailed description of the \textit{learning while recovering} algorithm.

\section{Technical Details}
\label{network_structure}

We present the details of architecture designs of the LVM and the EBM, as well as the hyper-parameters below. 

\textbf{Latent Variable Model:} The LVM $G_\alpha(X,h)$, using the ResNet50 \cite{ResHe2015} as an encoder backbone,
% takes image $X$ and latent variable $h$ as input, and
maps an image $X$ and a latent vector $h$ to a saliency map $\hat{Y}$. Specifically, we adopt the decoder from the MiDaS %depth estimation 
\cite{Ranftl2020} for its simplicity, which gradually aggregates the higher level features with lower level features via residual connections. We introduce the latent vector $h$ to the bottleneck of the LVM by concatenating the tiled $h$ with the highest level features of the encoder backbone, and then feed them to a $3\times3$ convolutional layer to obtain a feature map with the same size as the original highest level feature map of the encoder. The latent-vector-aware feature map is then fed to the decoder from \citet{Ranftl2020} to generate a final saliency map. As shown in Eq.~(\ref{equ:LVM_update}), the parameters of the LVM are updated with the revised predictions $\{\tilde{Y}\}$ provided by the EBM. Thus, immature $\{\tilde{Y}\}$ in the early stage of the cooperative learning might bring in fluctuation in training the LVM, which in turn affects the convergence of the MCMC samples $\{\tilde{Y}\}$.  
To stabilize the cooperative training, especially in the early stage, we let the LVM learn from not only  $\{\tilde{Y}\}$ but also $\{Y\}$. 
Specifically, we add an extra loss for the LVM as $\lambda\mathcal{L}_{ce}(G_\alpha(X,\tilde{h}),Y)$, where $\lambda$ linearly decreases to 0 during training, and $\mathcal{L}_{ce}$ is the cross-entropy loss.

\textbf{Energy-Based Model:}
The energy function $U_\theta(Y,X)$ is parameterized by a neural network that maps the channel-wise concatenation of $X$ and $Y$ to a scalar. Let \texttt{c\underline{k}s\underline{l}-\underline{n}} denote a \texttt{k}$\times$\texttt{k} Convolution-BatchNorm-ReLU layer with  \texttt{n} filters and a stride of \texttt{l}. Let \texttt{fc-\underline{n}} be a fully connected layer with  \texttt{n} filters. The $U_\theta(Y,X)$ is our framework consists of the following layers: \texttt{c3s1-32, c4s2-64, c4s2-128, c4s2-256, c4s1-1, fc-100}.

\begin{table*}[t!]
  \centering
  \scriptsize
  \renewcommand{\arraystretch}{0.9}
  \renewcommand{\tabcolsep}{0.4mm}
  \caption{Performance comparison with benchmark saliency prediction models, where \enquote{BkB} indicates the encoder backbone, and \enquote{R34} is ResNet34 backbone \cite{ResHe2015}, and \enquote{R50} is the ResNet50 backbone \cite{ResHe2015}.
  }
  \begin{tabular}{l|l|l|cccc|cccc|cccc|cccc|cccc|cccc}
  \hline
% \toprule
  &&&\multicolumn{4}{c|}{DUTS}&\multicolumn{4}{c|}{ECSSD}&\multicolumn{4}{c|}{DUT}&\multicolumn{4}{c|}{HKU-IS}&\multicolumn{4}{c|}{PASCAL-S}&\multicolumn{4}{c}{SOD} \\
    Method &Year&BkB& $S_{\alpha}\uparrow$&$F_{\beta}\uparrow$&$E_{\xi}\uparrow$&$\mathcal{M}\downarrow$& $S_{\alpha}\uparrow$&$F_{\beta}\uparrow$&$E_{\xi}\uparrow$&$\mathcal{M}\downarrow$& $S_{\alpha}\uparrow$&$F_{\beta}\uparrow$&$E_{\xi}\uparrow$&$\mathcal{M}\downarrow$& $S_{\alpha}\uparrow$&$F_{\beta}\uparrow$&$E_{\xi}\uparrow$&$\mathcal{M}\downarrow$& $S_{\alpha}\uparrow$&$F_{\beta}\uparrow$&$E_{\xi}\uparrow$&$\mathcal{M}\downarrow$& $S_{\alpha}\uparrow$&$F_{\beta}\uparrow$&$E_{\xi}\uparrow$&$\mathcal{M}\downarrow$ \\ \hline
  \multicolumn{27}{c}{Fully Supervised Models} \\ \hline
   PoolNet & 2019&R50& .887 & .840 & .910 & .037 & .919 & .913 & .938 & .038 & .831 & .748 & .848 & .054 & .919 & .903 & .945 & .030 & .865 & .835 & .896 & .065  & .820 & .804 & .834 & .084\\ 
   BASNet & 2019&R34& .876 & .823 & .896 & .048 & .910 & .913 & .938 & .040 & .836 & .767 & .865 & .057 & .909 & .903 & .943 & .032 & .838 & .818 & .879 & .076  & .798 & .792 & .827 & .094\\ 
   SCRN & 2019&R50& .885 & .833 & .900 & .040 & .920 & .910 & .933 & .041 & .837 & .749 & .847 & .056 & .916 & .894 & .935 & .034 & .869 & .833 & .892 & .063  & .817 & .790 & .829 & .087\\ 
   F3Net & 2020 &R50& .888 & .852 & .920 & .035 & .919 & .921 & .943 & .036 & .839 & .766 & .864 & .053 & .917 & .910 & .952 & .028 & .861 & .835 & .898 & .062  & .824 & .814 & .850 & .077\\ 
   ITSD & 2020&R50& .885 & .840 & .913 & .041 & .919 & .917 & .941 & .037 & .840 & .768 & .865 & .061 & .917 & .904 & .947 & .031 & .860 & .830 & .894 & .066  & .836 & .829 & .867 & .076\\ 
   LDF & 2020&R50& .892 & .861 & .925 & .034 & .919 & .923 & .943 & .036 & .839 & .770 & .865 & .052 & .920 & .913 & .953 & .028 & .842 & .768 & .863 & .064  & - & - & - & -\\ 
   UCNet+ & 2021&R50& .888 & .860 & .927 & .034 & .921 & .926 & .947 & .035 & .839 & .773 & .869 & .051 & .921 & .919 & .957 & \textbf{.026} & .851 & .825 & .886 & .069  & .828 & .827 & .856 & .076\\ 
  PAKRN &2021&R50& .900 & .876 & .935 & .033 & \textbf{.928} & .930 & .951 & .032 & .853 & .796 & .888 & .050 & .923 & \textbf{.919} & .955 & .028 & .858 & .838 & .896 & .067   &.833 & .836 & .866 & .074\\
  \textbf{Our\_F} &2021&R50& \textbf{.902} & \textbf{.877} & \textbf{.936} & \textbf{.032} & \textbf{.928} & \textbf{.935} & \textbf{.955} & \textbf{.030} & \textbf{.857} & \textbf{.798} & \textbf{.889} & \textbf{.049} & \textbf{.927} & .917 & \textbf{.960} & \textbf{.026} & \textbf{.873} & \textbf{.846} & \textbf{.909} & \textbf{.058}  & \textbf{.854} & \textbf{.850} & \textbf{.885} & \textbf{.064}\\\hline
  \multicolumn{27}{c}{Weakly Supervised Models} \\ \hline
   SSAL & 2020&R50& .803 & .747 & .865 & .062 & .863 & .865 & .908 & .061 & .785 & .702 & .835 & .068 & .865 & .858 & .923 & .047 & .798 & .773 & .854 & .093  & .750 & .743 & .801 & .108\\ 
  SCWS & 2021 &R50& .841 & \textbf{.818}& .901 & .049 & .879 & .894 & .924 & .051 & .813 & .751 & .856 & .060 & .883 & .892 & .938 & .038 & .821 & .815 & .877  & .078 & .782 & .791 & .833 &.090\\ 
%   \hline 
   \textbf{Our\_W} &2021&R50& \textbf{.847} & .816 & \textbf{.902} & \textbf{.048} & \textbf{.896} & \textbf{.896} & \textbf{.934} & \textbf{.045} & \textbf{.817} & \textbf{.762} & \textbf{.864} & \textbf{.058} & \textbf{.894} & \textbf{.893} & \textbf{.943} & \textbf{.037} & \textbf{.834} & \textbf{.823} & \textbf{.886} & \textbf{.073} & \textbf{.803} & \textbf{.793} & \textbf{.849}  & \textbf{.082} \\\hline
  \end{tabular}
  \label{tab:benchmark_model_comparison}
%   \vspace{-2mm}
\end{table*}

\begin{figure*}[!t]
% \vspace{-2mm}
   \begin{center}
   \begin{tabular}{c@{ } c@{ } c@{ } c@{ } c@{ } c@{ } c@{ } c@{ }c@{ }}
   {\includegraphics[width=0.101\linewidth]{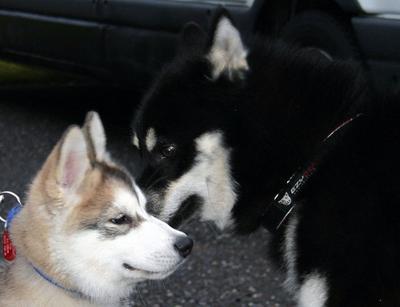}} &
        {\includegraphics[width=0.101\linewidth]{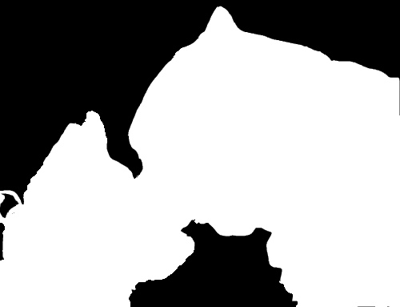}} &
        {\includegraphics[width=0.101\linewidth]{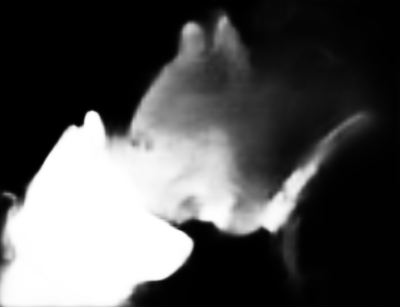}} &
        {\includegraphics[width=0.101\linewidth]{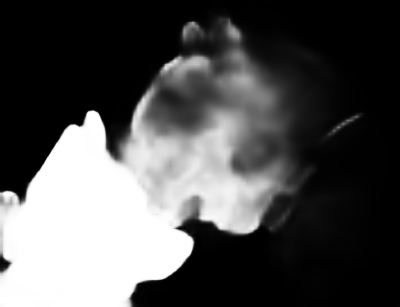}} & {\includegraphics[width=0.101\linewidth]{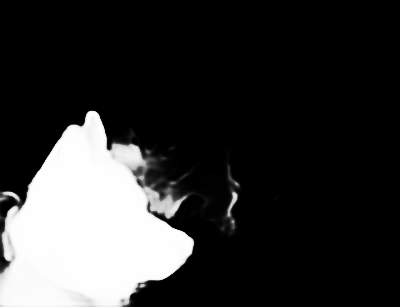}} &
        {\includegraphics[width=0.101\linewidth]{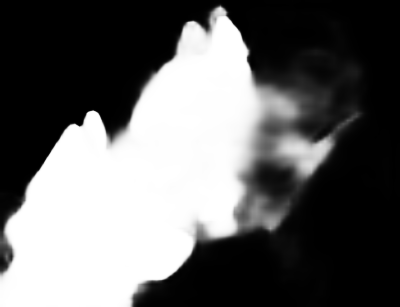}} &
        {\includegraphics[width=0.101\linewidth]{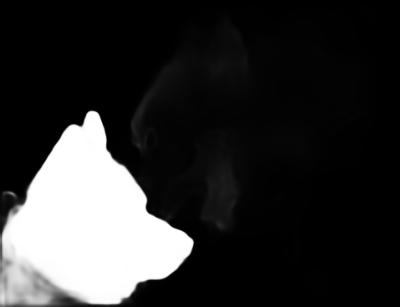}} &
        {\includegraphics[width=0.101\linewidth]{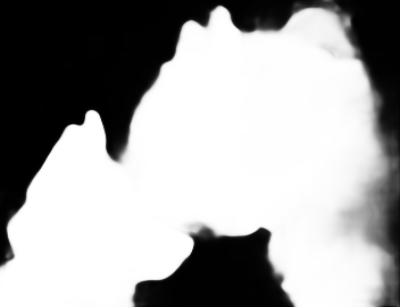}} &
        {\includegraphics[width=0.101\linewidth]{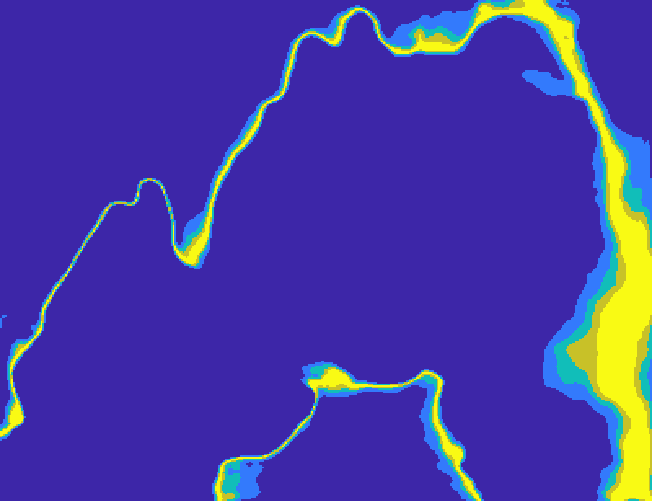}} \\
        {\includegraphics[width=0.101\linewidth]{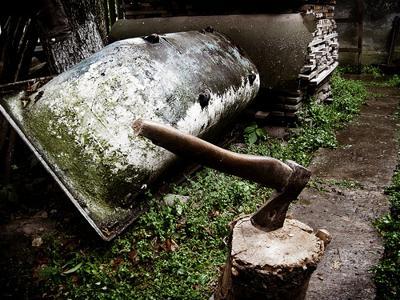}} &
        {\includegraphics[width=0.101\linewidth]{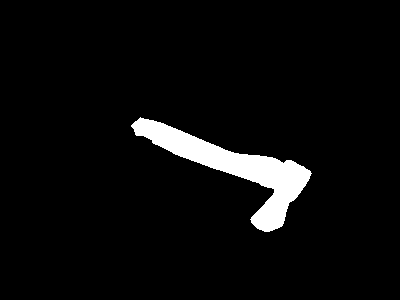}} &
        {\includegraphics[width=0.101\linewidth]{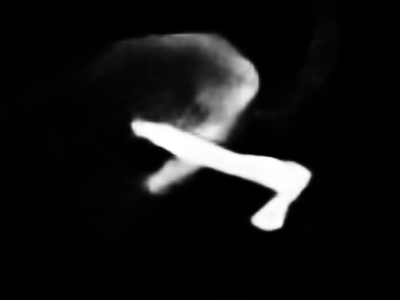}} &
        {\includegraphics[width=0.101\linewidth]{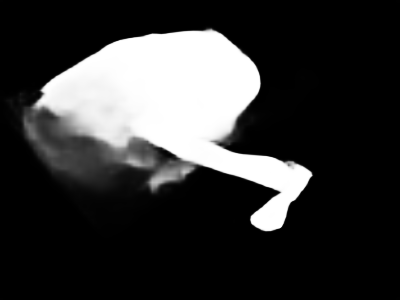}}& {\includegraphics[width=0.101\linewidth]{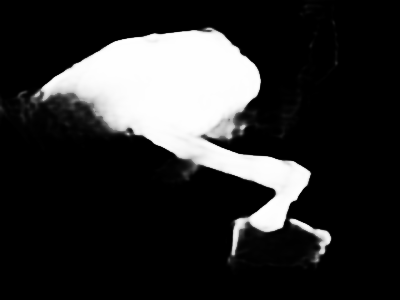}} &
        {\includegraphics[width=0.101\linewidth]{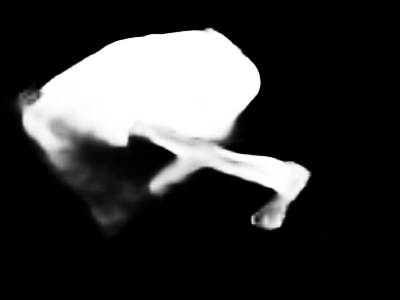}} &
        {\includegraphics[width=0.101\linewidth]{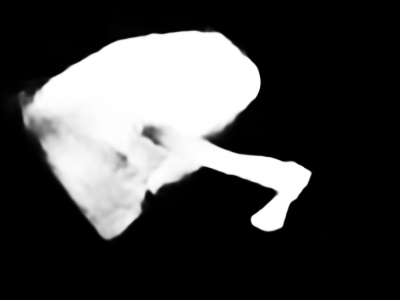}} &
        {\includegraphics[width=0.101\linewidth]{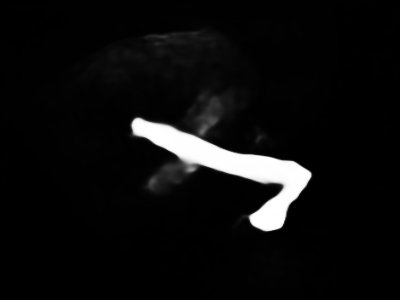}} &
        {\includegraphics[width=0.101\linewidth]{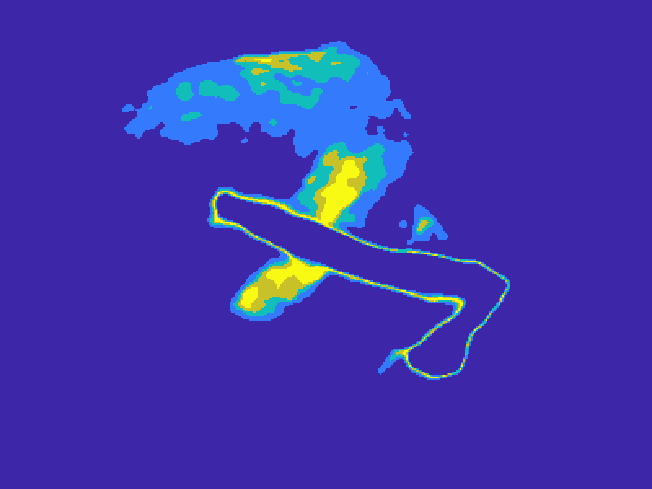}} \\
        {\includegraphics[width=0.101\linewidth]{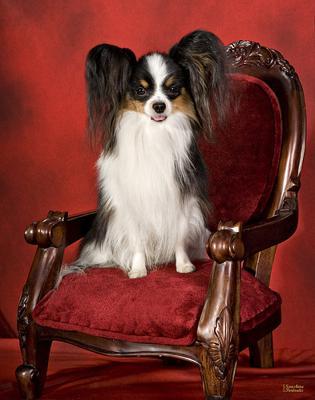}} &
        {\includegraphics[width=0.101\linewidth]{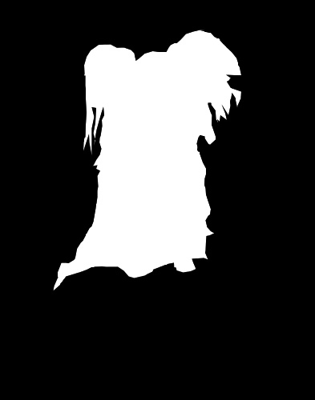}} &
        {\includegraphics[width=0.101\linewidth]{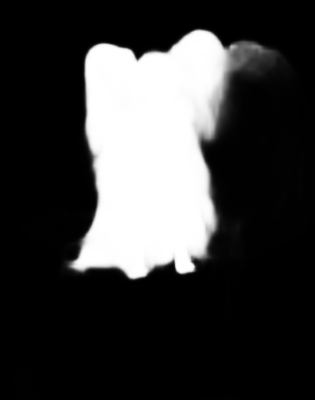}} &
        {\includegraphics[width=0.101\linewidth]{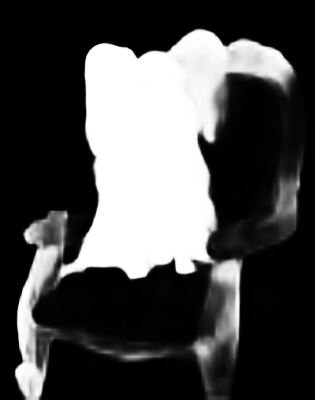}} & {\includegraphics[width=0.101\linewidth]{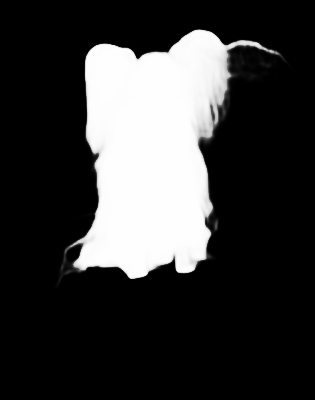}} &
        {\includegraphics[width=0.101\linewidth]{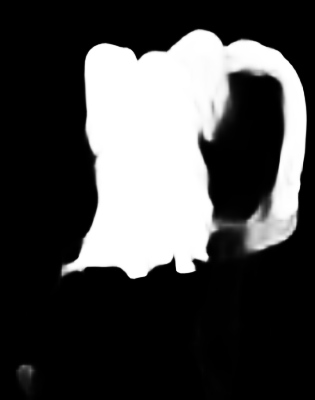}} &
        {\includegraphics[width=0.101\linewidth]{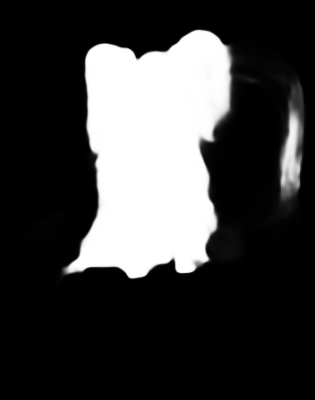}} &
        {\includegraphics[width=0.101\linewidth]{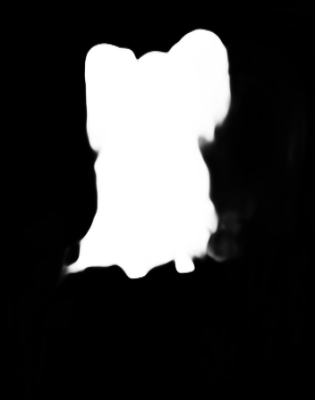}} &
        {\includegraphics[width=0.101\linewidth]{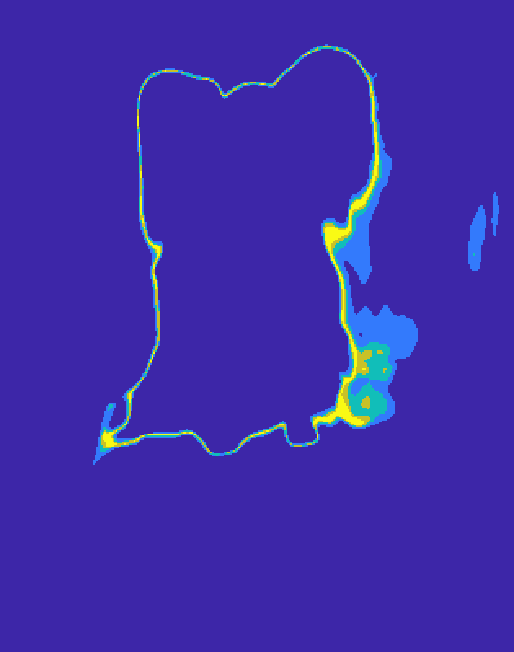}} \\
        {\includegraphics[width=0.101\linewidth]{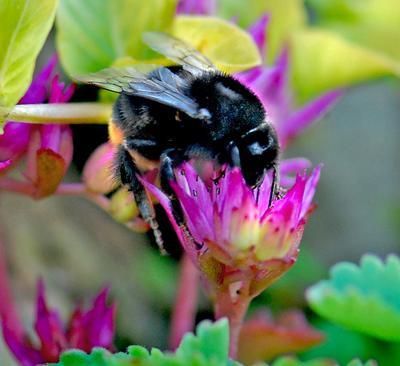}} &
        {\includegraphics[width=0.101\linewidth]{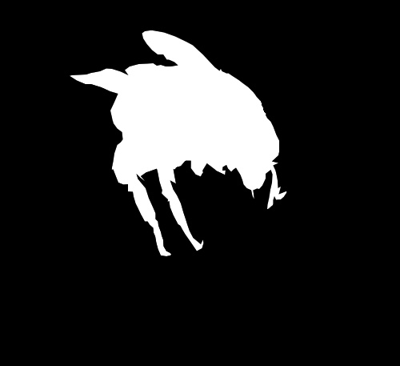}} &
        {\includegraphics[width=0.101\linewidth]{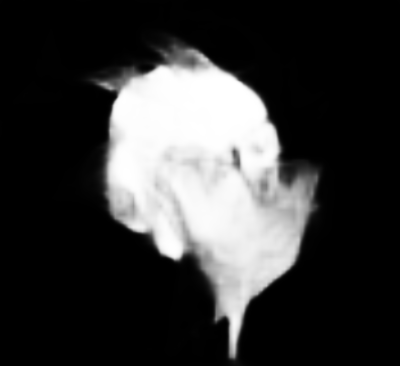}} &
        {\includegraphics[width=0.101\linewidth]{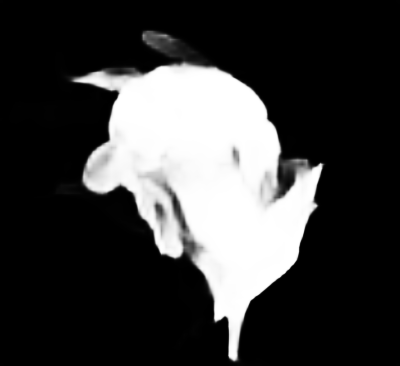}} & {\includegraphics[width=0.101\linewidth]{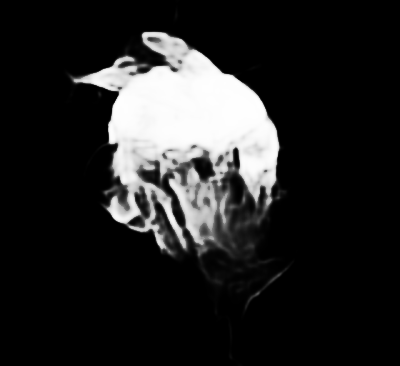}} &
        {\includegraphics[width=0.101\linewidth]{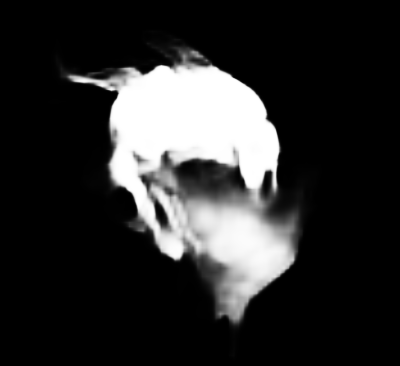}} &
        {\includegraphics[width=0.101\linewidth]{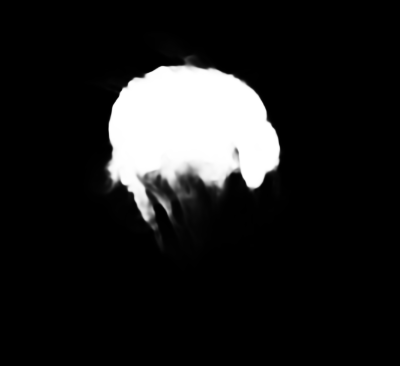}} &
        {\includegraphics[width=0.101\linewidth]{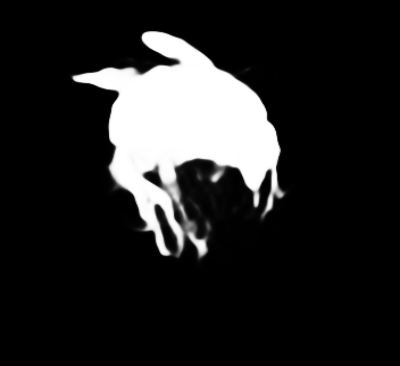}} &
        {\includegraphics[width=0.101\linewidth]{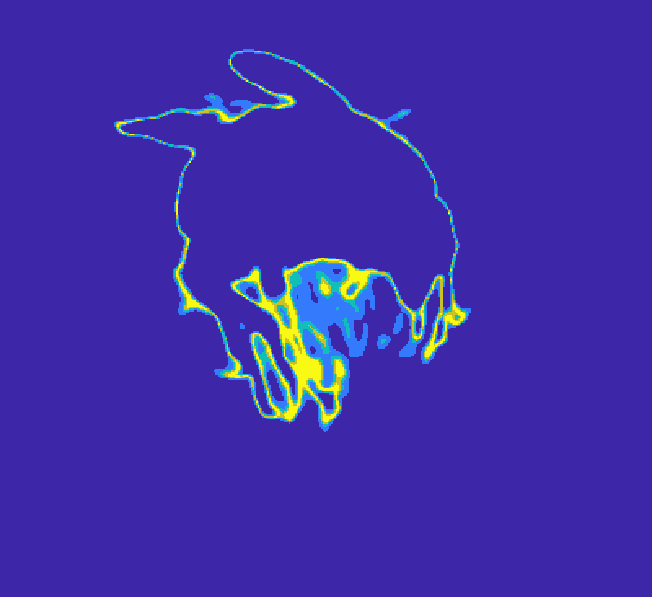}} \\
        {\includegraphics[width=0.101\linewidth]{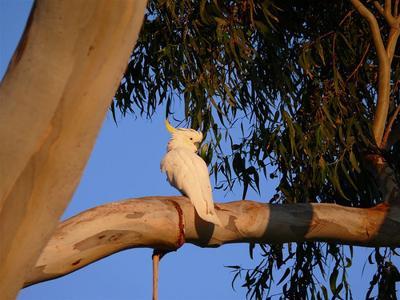}} &
        {\includegraphics[width=0.101\linewidth]{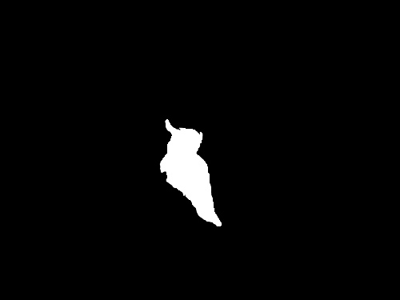}} &
        {\includegraphics[width=0.101\linewidth]{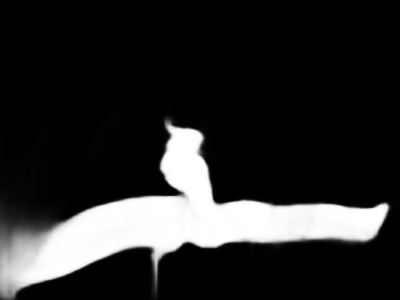}} &
        {\includegraphics[width=0.101\linewidth]{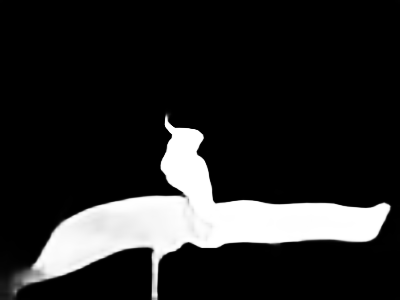}} & {\includegraphics[width=0.101\linewidth]{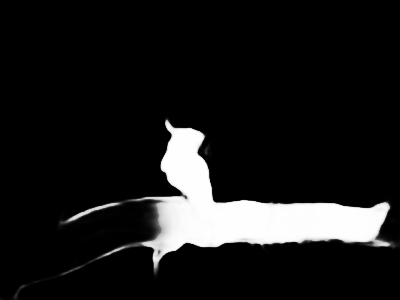}} &
        {\includegraphics[width=0.101\linewidth]{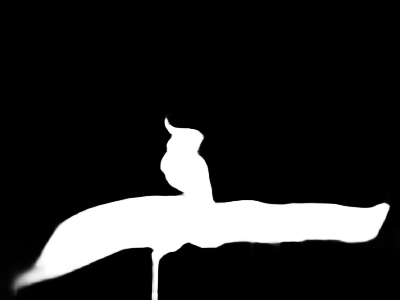}} &
        {\includegraphics[width=0.101\linewidth]{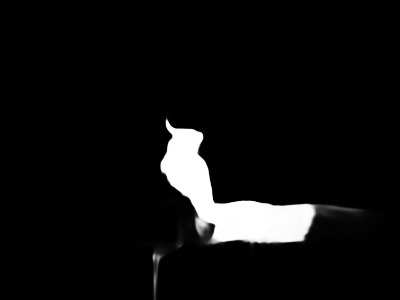}} &
        {\includegraphics[width=0.101\linewidth]{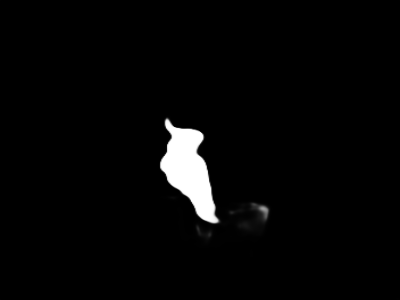}} &
        {\includegraphics[width=0.101\linewidth]{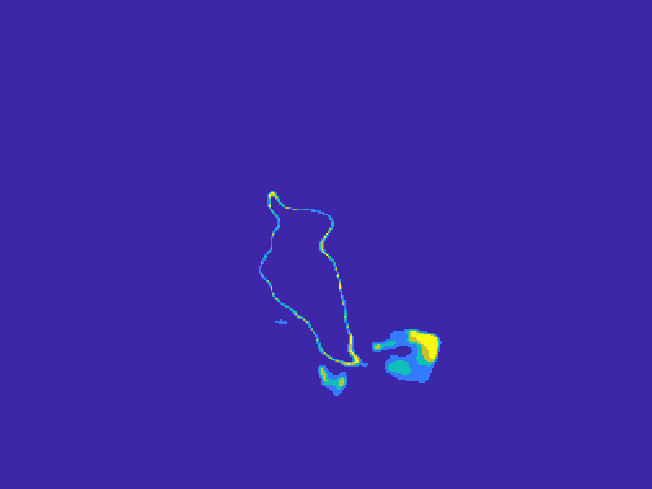}} \\
        \multicolumn{1}{c}{\footnotesize{Image}}
        &\multicolumn{1}{c}{\footnotesize{GT}}
        &\multicolumn{1}{c}{\footnotesize{SCRN}}
        &\multicolumn{1}{c}{\footnotesize{F3Net}}
        &\multicolumn{1}{c}{\footnotesize{ITSD}}
        &\multicolumn{1}{c}{\footnotesize{LDF}}
        &\multicolumn{1}{c}{\footnotesize{PAKRN}}
        &\multicolumn{1}{c}{\footnotesize{Ours\_F}}
        &\multicolumn{1}{c}{\footnotesize{Uncertainty}}\\
   \end{tabular}
   \end{center}
%   \vspace{-2mm}
\caption{\small Comparison of qualitative results of different fully supervised saliency prediction models.}
\vspace{-2mm}
   \label{fig:visual_comparison}
\end{figure*}

\textbf{Implementation Details:}
We train our model with a maximum of 30 epochs.
Each image is rescaled to $352\times 352$. We set the number of dimensions of the latent space as $d=8$. The number of Langevin steps is $K=5$ and the Langevin step sizes for EBM and LVM are 0.4 and 0.1.
The learning rates of the LVM and EBM are initialized to $5 \times 10^{-5}$ 
and $10^{-3}$ %0.001%1e-3 
respectively. We use Adam optimizer with momentum 0.9 and decrease the learning rates by 10\% after every 20 epochs. 
It takes 20 hours to train the model with a batch size of 7 using a single NVIDIA GeForce RTX 2080Ti GPU.

\section{Experiments}
We conduct~a~series of experiments to~test~the~performances of the proposed generative cooperative frameworks for saliency prediction. We start from~experiment~setup.

\textbf{Datasets:}
We use the DUTS dataset~\cite{imagesaliency} to train the fully supervised model, and S-DUTS \cite{jing2020weakly} dataset with scribble annotations to train the weakly supervised model. Testing images include (1) DUTS testing dataset, (2) ECSSD \cite{Hierarchical:CVPR-2013}, (3) DUT \cite{Manifold-Ranking:CVPR-2013}, (4) HKU-IS \cite{MDF:CVPR-2015}, (5) PASCAL-S \cite{pascal_s_dataset} and (6)SOD dataset \cite{sod_dataset}.

\textbf{Compared methods:} We compare our method against state-of-the-art fully supervised saliency detection methods, \eg, PoolNet \cite{Liu2019PoolSal}, BASNet \cite{BASNet_Sal}, SCRN \cite{SCRN_iccv}, F3Net \cite{F3Net_aaai2020}, ITSD \cite{itsd_sal}, LDF \cite{wei2020label}, UCNet+ \cite{ucnet_jornal} and PAKRN \cite{xu2021locate}. UCNet+ \cite{ucnet_jornal} is the only generative framework. We also compare our weakly supervised solution with the scribble saliency detection models, \eg, SSAL \cite{jing2020weakly} and SCWS \cite{structure_consistency_scribble}.

\textbf{Evaluation Metrics:} We evaluate performance of our models and compared methods with four saliency evaluation metrics, including Mean Absolute Error ($\mathcal{M}$), mean F-measure ($F_{\beta}$), mean E-measure ($E_{\xi}$) \cite{Fan2018Enhanced} and S-measure ($S_{\alpha}$) \cite{fan2017structure}.

\begin{table*}[t!]
  \centering
  \scriptsize
  \renewcommand{\arraystretch}{1.0}
  \renewcommand{\tabcolsep}{0.55mm}
  \caption{Experimental results of using EBMs as refinement modules and ablation study.
  }
  \begin{tabular}{l|cccc|cccc|cccc|cccc|cccc|cccc}
  \hline
%   \toprule
  &\multicolumn{4}{c|}{DUTS}&\multicolumn{4}{c|}{ECSSD}&\multicolumn{4}{c|}{DUT}&\multicolumn{4}{c|}{HKU-IS}&\multicolumn{4}{c|}{PASCAL-S}&\multicolumn{4}{c}{SOD} \\
     & $S_{\alpha}\uparrow$&$F_{\beta}\uparrow$&$E_{\xi}\uparrow$&$\mathcal{M}\downarrow$& $S_{\alpha}\uparrow$&$F_{\beta}\uparrow$&$E_{\xi}\uparrow$&$\mathcal{M}\downarrow$& $S_{\alpha}\uparrow$&$F_{\beta}\uparrow$&$E_{\xi}\uparrow$&$\mathcal{M}\downarrow$& $S_{\alpha}\uparrow$&$F_{\beta}\uparrow$&$E_{\xi}\uparrow$&$\mathcal{M}\downarrow$& $S_{\alpha}\uparrow$&$F_{\beta}\uparrow$&$E_{\xi}\uparrow$&$\mathcal{M}\downarrow$& $S_{\alpha}\uparrow$&$F_{\beta}\uparrow$&$E_{\xi}\uparrow$&$\mathcal{M}\downarrow$ \\
  \hline
  \multicolumn{25}{c}{EBM as Refinement Module} \\ \hline
  BASN\_R & .891 & .842 & .889 & .041 & .926 & .921 & .947 & .035 & .839 & .781 & .870 & .051 & .919 & .925 & .942 & .031 & .837 & .749 & .857 & .070 & .852 & .859 & .869 & .090\\
  SCRN\_R & .899 & .857 & .923 & .034 & .920 & .921 & .938 & .037 & .831 & .748 & .854 & .053 & .921 & .919 & .958 & .027 & .857 & .769 & .871 & .062 & .857 & .867 & .873 & .080\\
  \hline
  \multicolumn{25}{c}{Ablation Study} \\ \hline
   $G_\alpha(X)$ & .878 & .835 & .918 & .038 & .916& .915 & .946 & .036 & .826 & .751 & .862 & .058 & .912 & .901 & .952 & .030 & .856 & .830 & .899 & .064  & .829 & .827 & .871 & .072\\ 
   $G_\alpha(X,h)$ & .897 & .858 & .932 & .034 & .918 & .923 & .946 & .034 & .837 & .777 & .882& .051 & .914 & .913 & .957 & .028 & .863 & .835 & .900 & .062  & .831 & .830 & .874 & .070\\
%   \hline
   ITSD & .885 & .840 & .913 & .041 & .919 & .917 & .941 & .037 & .840 & .768 & .865 & .061 & .917 & .904 & .947 & .031 & .860 & .830 & .894 & .066  & .836 & .829 & .867 & .076\\ 
   ITSD\_Ours & .914 & .880 & .945 & .030 & .938 & .935 & .959 & .029 & .860 & .803 & .901 & .044 & .933 & .927 & .971 & .026 & .875 & .848 & .921 & .055  & .845 & .835 & .880 & .067\\
   VGG16\_Ours & .906 & .876 & .941 & .032 & .939 & .933 & .953 & .030 & .857 & .799 & .893 & .048 & .929 & .923 & .959 & .027 & .871 & .844 & .907 & .058  & .841 & .838 & .871 & .066\\ 
   Our\_F & .902 & .877 & .936 & .032 & .928 & .935 & .955 & .030 & .857 & .798 & .889 & .049 & .927 & .917 & .960 & .026 & .873 & .846 & .909 & .058  & .854 & .850 & .885 & .064\\ \hline
  \end{tabular}
  \label{tab:ablation_study}
  \vspace{-2mm}
\end{table*}

% \noindent\textbf{Parameter Updating}
\begin{figure}[!t]
   \begin{center}
   \begin{tabular}{c@{ } c@{ } c@{ } c@{ }}
        {\includegraphics[width=0.235\linewidth]{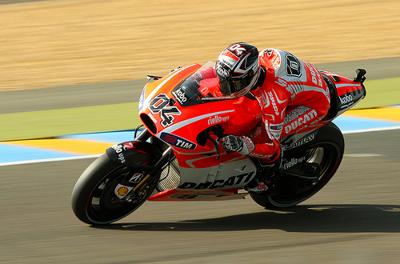}} &
        {\includegraphics[width=0.235\linewidth]{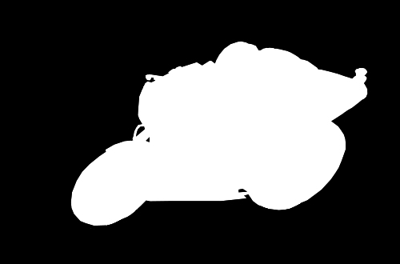}} &
        {\includegraphics[width=0.235\linewidth]{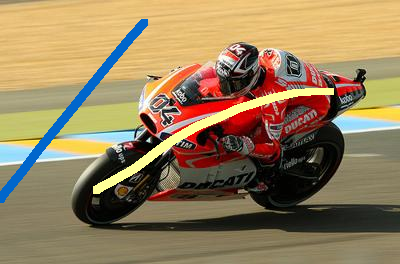}} &
        {\includegraphics[width=0.235\linewidth]{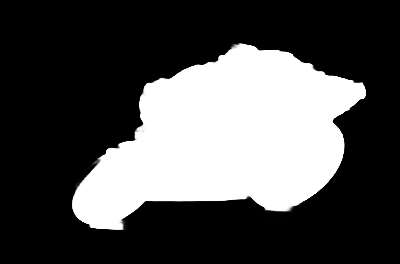}}\\
        {\includegraphics[width=0.235\linewidth]{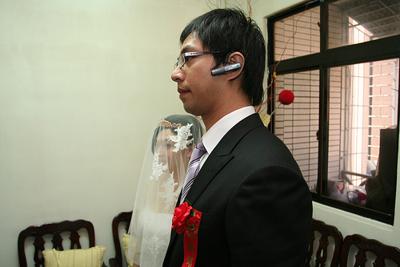}} &
        {\includegraphics[width=0.235\linewidth]{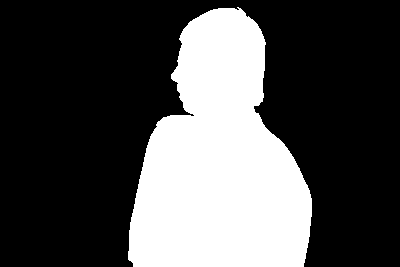}} &
        {\includegraphics[width=0.235\linewidth]{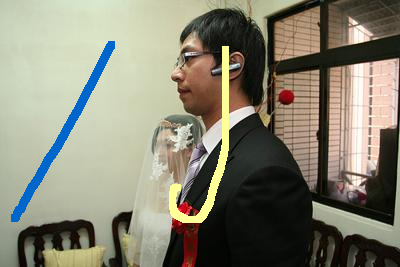}} &
        {\includegraphics[width=0.235\linewidth]{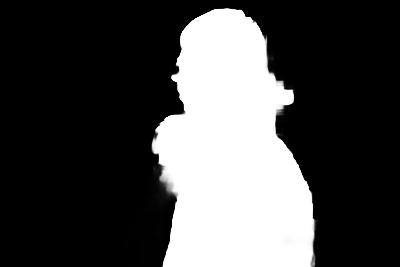}}\\
         {\includegraphics[width=0.235\linewidth]{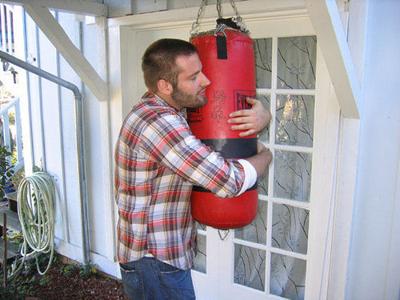}} &
        {\includegraphics[width=0.235\linewidth]{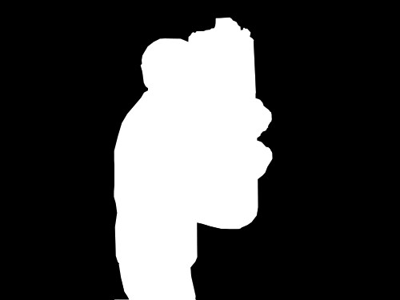}} &
        {\includegraphics[width=0.235\linewidth]{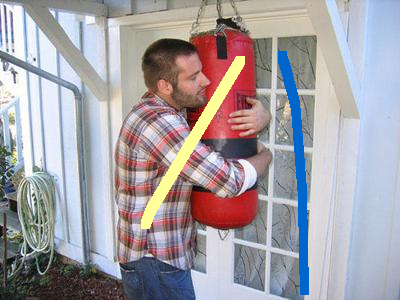}} &
        {\includegraphics[width=0.235\linewidth]{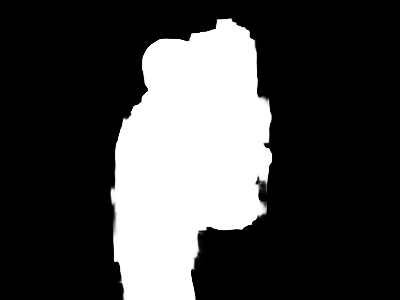}}\\
        \footnotesize{Image} & \footnotesize{GT} & \footnotesize{Scribble} &  \footnotesize{Recovered}\\
   \end{tabular}
   \end{center}
%   \vspace{-2mm}
\caption{\small Learning from images with scribble annotations. Each row shows one example of annotation recovery during training. The columns from left to right present training input images (\enquote{images}), ground truth annotations (\enquote{GT}) that are unknown for the learning algorithm, scribble annotations (\enquote{scribble}) as weak labels for training, and the output recovered annotations (\enquote{Recovered}) using the proposed \enquote{cooperative learning while recovering} strategy. 
}
% \vspace{-2mm}
   \label{fig:scribble_annotation}
\end{figure}

\subsection{Fully Supervised Saliency Prediction}

We first test the performance of our fully supervised generative cooperative saliency prediction framework. 

\textbf{Quantitative comparison:}
We compare the performance of our models and the compared methods in Table \ref{tab:benchmark_model_comparison}, where \enquote{Ours\_F} denotes the proposed fully supervised models. We observe consistent performance improvement of \enquote{Ours\_F} over six testing datasets compared with benchmark models, which clearly shows the advantage of our model. Note that, we adopt an existing decoder structure, \ie, MiDaS decoder \cite{Ranftl2020}, for the latent variable model in our proposed framework due to its easy implementation. We conduct an ablation study in Section \ref{ablation} to further investigate the design of the decoder. Since our model uses a stochastic method, \ie, cooperative sampling, for prediction, we report the mean prediction to evaluate the performance of our models. Also, we observe relatively stable performance for different samples of predictions in larger testing datasets, \eg,~DUTS testing dataset \cite{imagesaliency}, and slightly fluctuant performance in smaller testing datasets, \eg,~SOD \cite{sod_dataset} testing dataset.

\begin{figure}[!t]
   \begin{center}
   \begin{tabular}{c@{ } c@{ } c@{ } c@{ } c@{ } c@{ } }
      {\includegraphics[width=0.15\linewidth]{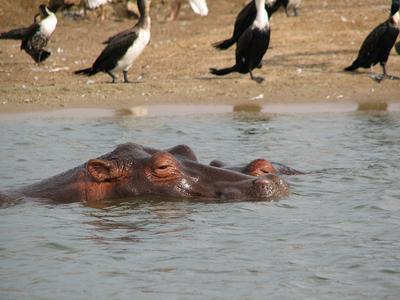}} &
        {\includegraphics[width=0.15\linewidth]{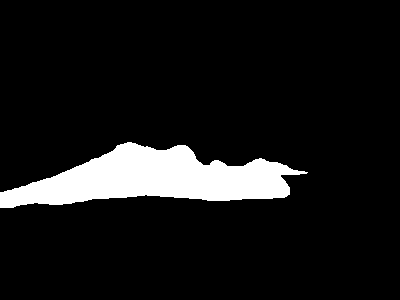}} &
        {\includegraphics[width=0.15\linewidth]{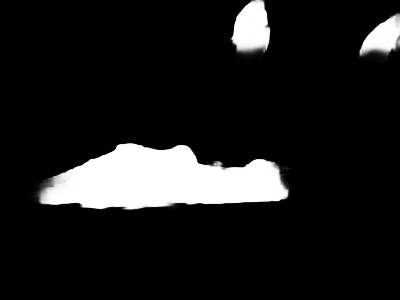}} &
        {\includegraphics[width=0.15\linewidth]{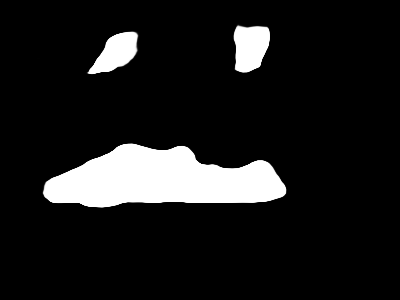}} &
        {\includegraphics[width=0.15\linewidth]{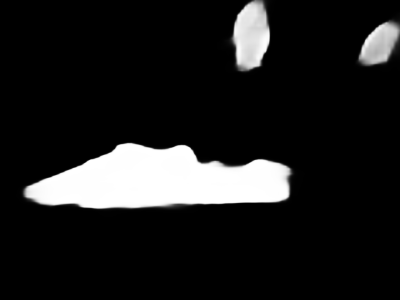}} &
        {\includegraphics[width=0.15\linewidth]{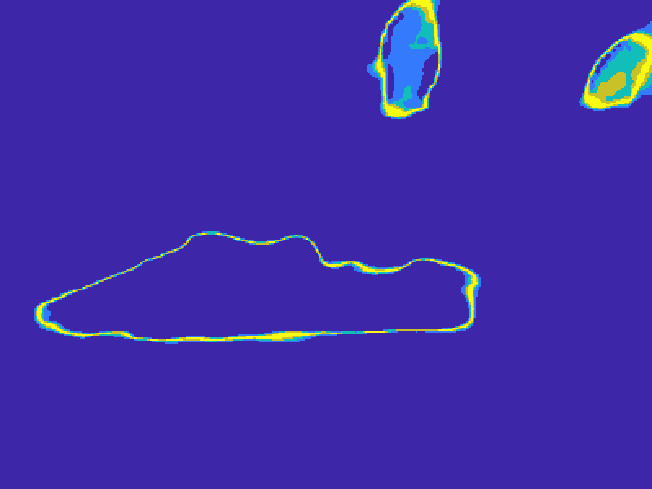}} \\
        {\includegraphics[width=0.15\linewidth]{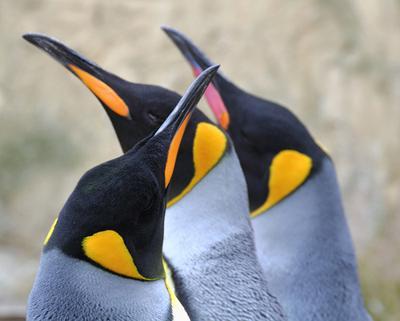}} &
        {\includegraphics[width=0.15\linewidth]{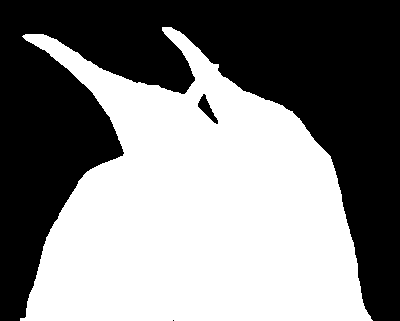}} &
        {\includegraphics[width=0.15\linewidth]{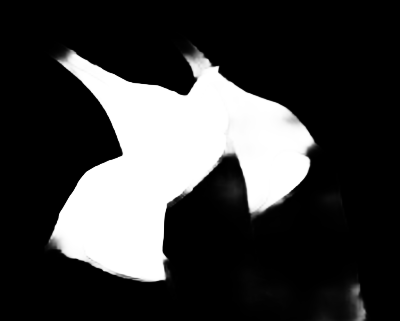}} &
        {\includegraphics[width=0.15\linewidth]{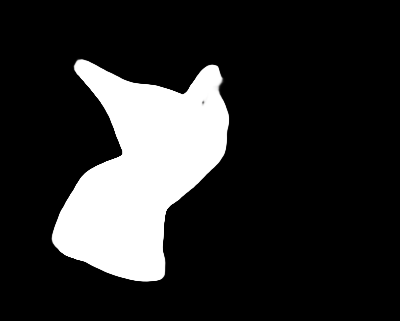}} &
        {\includegraphics[width=0.15\linewidth]{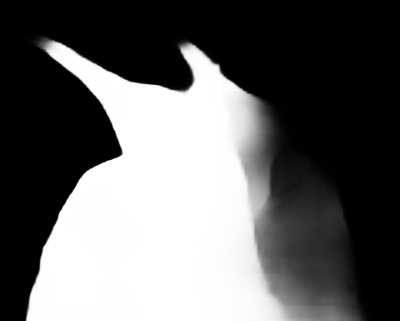}} &
        {\includegraphics[width=0.15\linewidth]{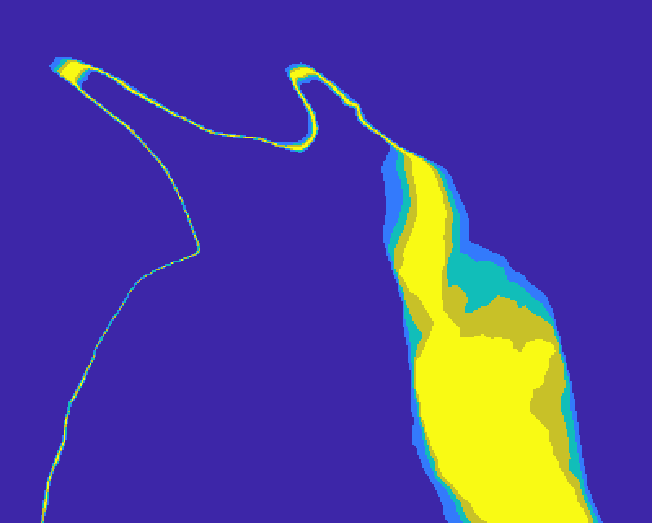}} \\
   {\includegraphics[width=0.15\linewidth]{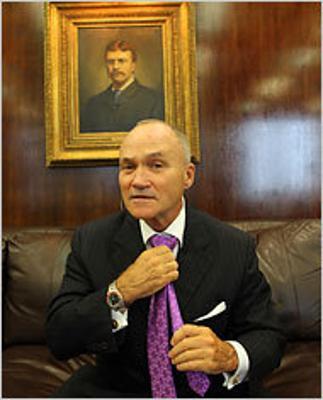}} &
        {\includegraphics[width=0.15\linewidth]{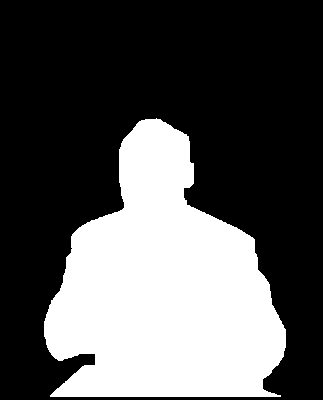}} &
        {\includegraphics[width=0.15\linewidth]{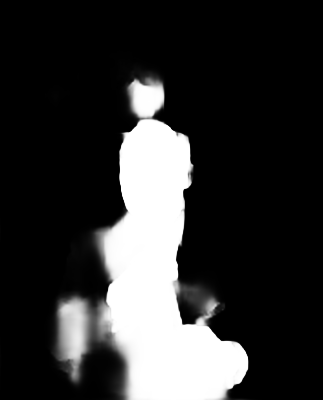}} &
        {\includegraphics[width=0.15\linewidth]{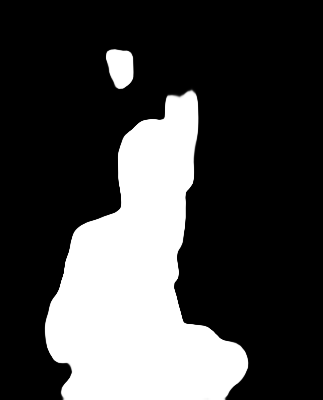}} &
        {\includegraphics[width=0.15\linewidth]{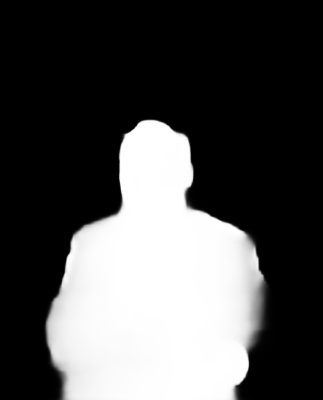}} &
        {\includegraphics[width=0.15\linewidth]{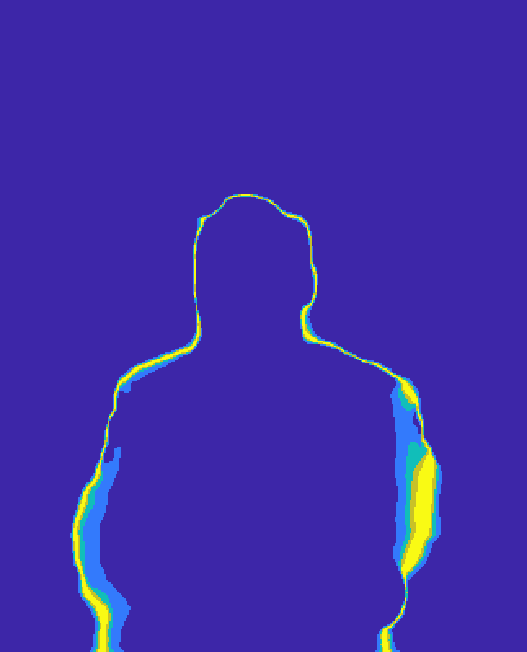}} \\
        {\includegraphics[width=0.15\linewidth]{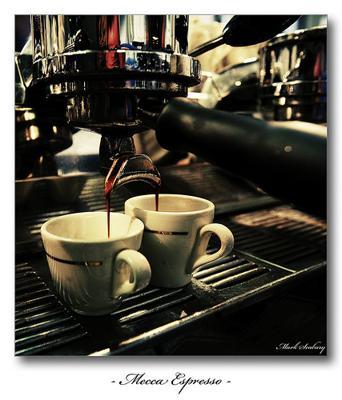}} &
        {\includegraphics[width=0.15\linewidth]{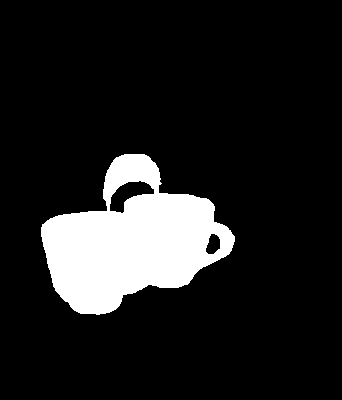}} &
        {\includegraphics[width=0.15\linewidth]{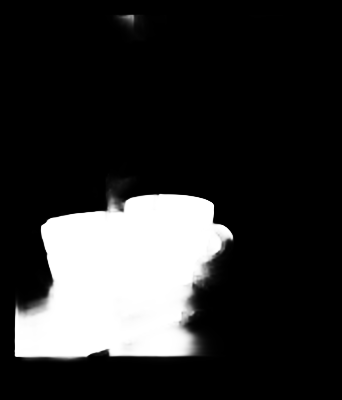}} &
        {\includegraphics[width=0.15\linewidth]{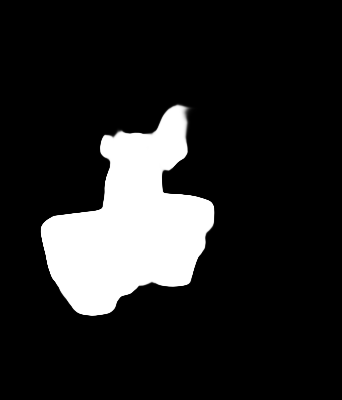}} &
        {\includegraphics[width=0.15\linewidth]{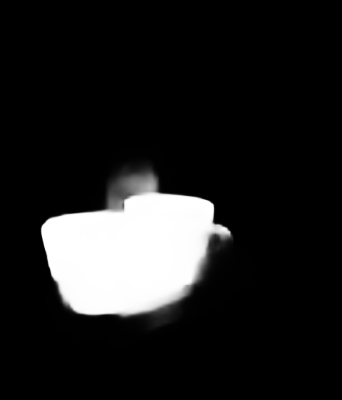}} &
        {\includegraphics[width=0.15\linewidth]{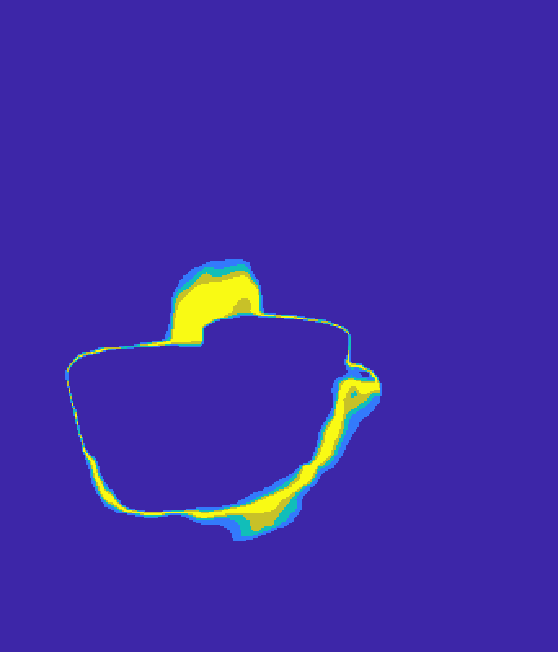}} \\
        
        \multicolumn{1}{c}{\footnotesize{Image}} & \multicolumn{1}{c}{\footnotesize{GT}} &
        \multicolumn{1}{c}{\footnotesize{SSAL}}& \multicolumn{1}{c}{\footnotesize{SCWS}}&
        \multicolumn{1}{c}{\footnotesize{Ours\_W}} & 
        \multicolumn{1}{c}{\footnotesize{Uncer.}}
        \\
   \end{tabular}
   \end{center}
%   \vspace{-2mm}
\caption{\small Comparison of qualitative results obtained by different weakly supervised models learned from scribble annotations. Each row of example illustrates an input testing image, the corresponding ground truth saliency map as reference, the predictions from compared methods SSAL and SCWS, the mean prediction (\enquote{Ours\_W}) and the uncertainty map (\enquote{Uncer.}) of our method .   
}
\vspace{-2mm}
   \label{fig:visual_comparison_weak}
\end{figure}

\textbf{Qualitative comparison:}
Figure~\ref{fig:visual_comparison} displays some qualitative results of the saliency predictions produced by our method and the compared methods. Each row corresponds to one example, and shows an input testing image, the corresponding ground truth saliency map, saliency maps predicted by SCRN, F3Net, ITSD, LDF and PAKRN, followed by the mean predicted saliency map and the pixel-wise uncertainty map of our model. The uncertainty map indicates the model confidence in predicting saliency from a given image, and is computed as the entropy \cite{kendall2015bayesian,what_uncertainty,ucnet_jornal,jing2020uc} of our model predictions. Results show that our method can not only produce visually reasonable saliency maps for input images 
but also meaningful uncertainty maps that are consistent with human perception.

\textbf{Prediction time and model size comparison:}
We have two main modules in our framework, namely a latent variable model and an energy-based model. The former takes the ResNet50 \cite{ResHe2015} backbone as encoder, and the MiDaS \cite{Ranftl2020} decoder for feature aggregation, leading to a model parameter size of 55M for the LVM. The latter adds 1M extra parameters to the cooperative learning framework. Thus, our model size is a total of 56M, which is comparable with mainstream saliency detection models, \eg, F3Net \cite{F3Net_aaai2020} has 48M parameters. As to the cooperative prediction time, it costs approximately 0.08 seconds to output a single prediction of saliency map, which is comparable with existing solutions as well.

\subsection{Weakly Supervised Saliency Prediction}
We then evaluate our weakly supervised generative cooperative saliency prediction framework on a dataset with scribble annotations \cite{jing2020weakly}, and show prediction performance on six testing sets in Table \ref{tab:benchmark_model_comparison}, where \enquote{Our\_W} in the \enquote{Weakly Supervised Models} panel denotes our model. Figure \ref{fig:scribble_annotation} shows some examples of annotation recovery during training, where each row of example displays an input training image, the ground truth annotation as reference, scribble annotation used for training (the yellow scribble indicates the salient region, and the blue scribble indicates the background region), and the recovered saliency annotation obtained by our method. 
We compare our model with baseline methods, \eg, SSAL \cite{jing2020weakly} and SCWS \cite{structure_consistency_scribble}. The better performance of our model in testing shows the effectiveness of the proposed ``cooperative learning while recovering'' algorithm. 
Figure~\ref{fig:visual_comparison_weak} displays a comparison of qualitative results obtained by different weakly supervised saliency prediction methods in testing.

\subsection{Energy Function as a Refinement Module}
\label{ebm_refine_subsection}
\begin{figure}[!t]
  \begin{center}
  \begin{tabular}{c@{ } c@{ } c@{ } c@{ } c@{ } c@{ }}
        {\includegraphics[width=0.152\linewidth]{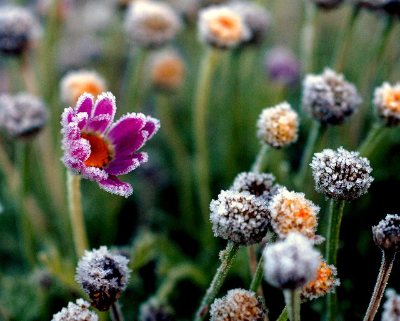}} &
        {\includegraphics[width=0.152\linewidth]{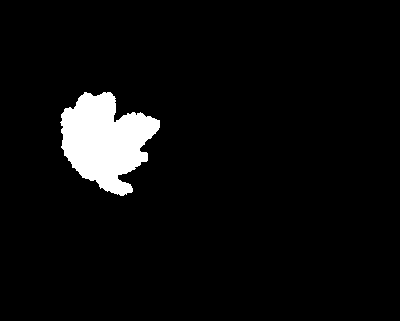}}
                &
        {\includegraphics[width=0.152\linewidth]{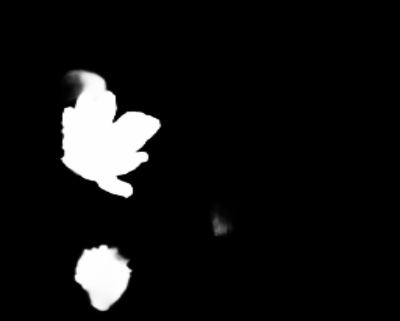}}&
        {\includegraphics[width=0.152\linewidth]{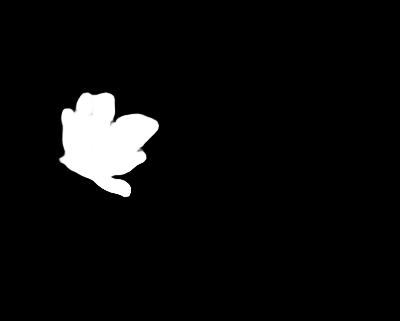}}&
        {\includegraphics[width=0.152\linewidth]{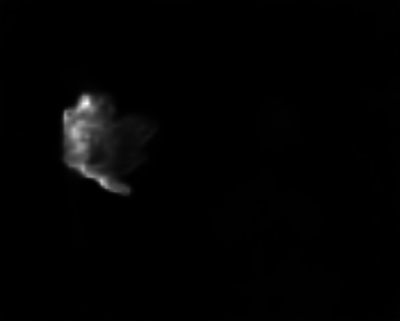}}&
        {\includegraphics[width=0.152\linewidth]{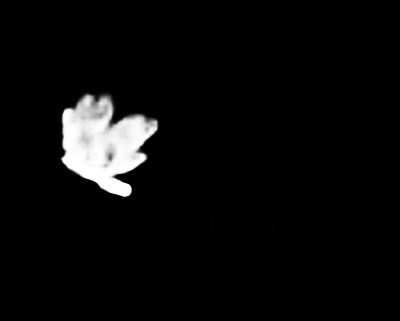}}
        \\
        {\includegraphics[width=0.152\linewidth]{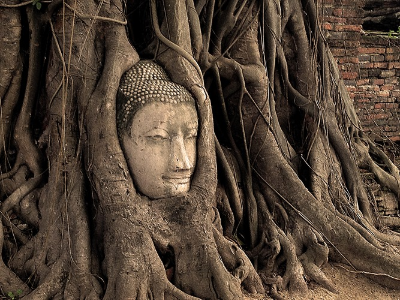}} &
        {\includegraphics[width=0.152\linewidth]{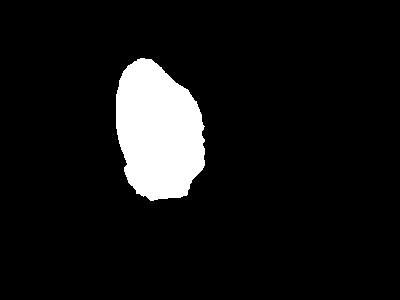}} &
        {\includegraphics[width=0.152\linewidth]{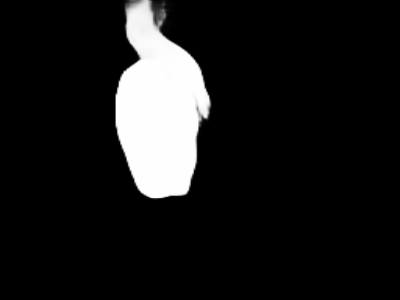}}&
        {\includegraphics[width=0.152\linewidth]{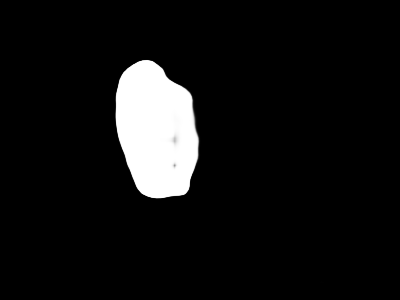}}&
        {\includegraphics[width=0.152\linewidth]{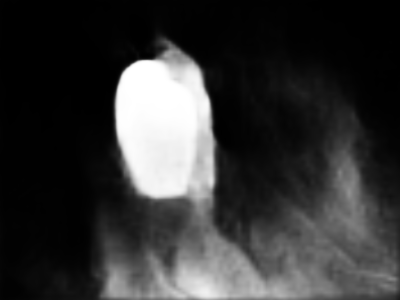}}&
        {\includegraphics[width=0.152\linewidth]{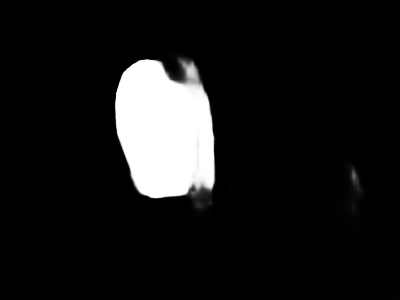}}
        \\
         {\includegraphics[width=0.152\linewidth]{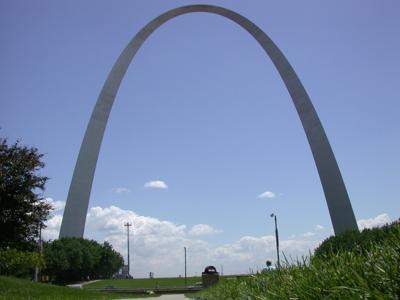}} &
        {\includegraphics[width=0.152\linewidth]{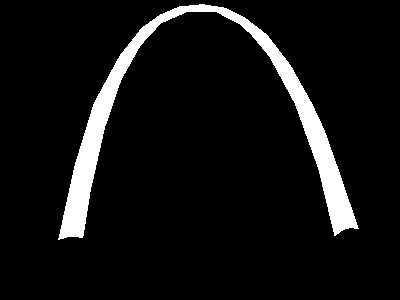}} &
                {\includegraphics[width=0.152\linewidth]{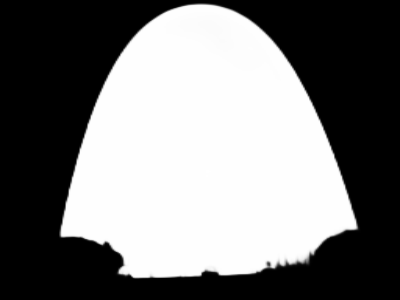}}&
        {\includegraphics[width=0.152\linewidth]{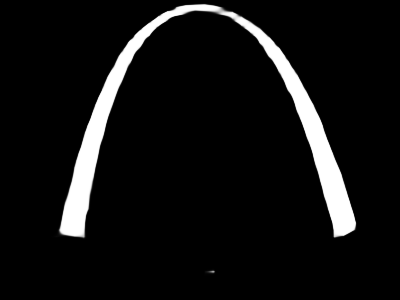}}&
        {\includegraphics[width=0.152\linewidth]{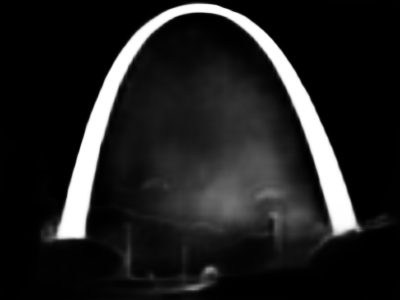}}&
        {\includegraphics[width=0.152\linewidth]{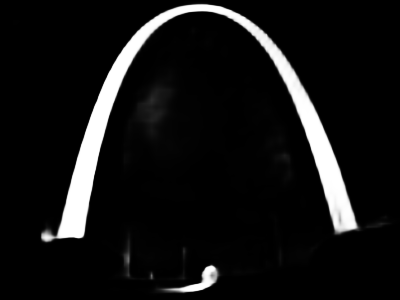}}
        \\
        \footnotesize{Image} & \footnotesize{GT} 
        & \footnotesize{BASN}
        & \footnotesize{BASN\_R}
        & \footnotesize{SCRN}
        & \footnotesize{SCRN\_R} \\
  \end{tabular}
  \end{center}
%   \vspace{-2mm}
\caption{\small Comparison of qualitative results of base models without and with the trained EBM as a refinement module. \enquote{BASN} and \enquote{SCRN} are base models. \enquote{BASN\_R} and \enquote{SCRN\_R} are the corresponding methods with EBM refinement.}
\vspace{-2mm}
  \label{fig:visual_ebm_refinement}
\end{figure}

As shown in Eq.~(\ref{equ:ebm_Langevin}), the EBM can iteratively refine the saliency prediction by Langevin sampling. With a well-trained energy function, we can treat it as a refinement module to refine predictions from existing saliency detection models. To demonstrate this idea, we select \enquote{BASN} \cite{BASNet_Sal} and \enquote{SCRN} \cite{SCRN_iccv} as base models due to the accessibility of their codes and predictions. We refine their predictions with the trained EBM and denote them by \enquote{BASN\_R} and \enquote{SCRN\_R}, respectively. Performances are shown in Table \ref{tab:ablation_study}.  
Comparing with the performance of the base models in Table \ref{tab:benchmark_model_comparison}, we observe consistent performance improvements by using the trained EBM for refinement in Table \ref{tab:ablation_study}. 
We show three examples of these models with and without EBM refinement in Figure~\ref{fig:visual_ebm_refinement}.~The qualitative improvement due to the usage of the EBM refinement verifies the usefulness of learned energy~function.

\subsection{Ablation Study}
\label{ablation}
We conduct the following experiments as shown in Table \ref{tab:ablation_study} to further analyze our proposed framework.
\textbf{Training a deterministic noise-free encoder-decoder $G_\alpha$:
% deterministic latent variable model:
} We remove the latent vector $h$ from our noise-injected encoder-decoder $G_\alpha(X,h)$ and obtain a deterministic noise-free encoder-decoder $G_\alpha(X)$. We train $G_\alpha(X)$ with the cross-entropy loss as in those conventional deterministic saliency prediction models. 
In comparison with the state-of-the-art deterministic saliency detection model, \ie, PAKRN \cite{xu2021locate}, 
$G_\alpha(X)$ shows inferior performance due to its usage of a relatively small decoder \cite{Ranftl2020}.
However, the superior performance of \enquote{Ours\_F}, which is built upon $G_\alpha(X,h)$ that shares the same decoder structure with $G_\alpha(X)$, has exhibited the usefulness of the latent vector $h$ for generative modeling and verified the effectiveness of the EBM for cooperative learning.

\textbf{Training a latent variable model $G_\alpha(X,h)$ without the EBM:}
% without the \enquote{Langevin sampler} in Eq. \ref{equ:ebm_Langevin}.
To further validate the importance of the cooperative training, we train a single latent variable model $G_\alpha(X,h)$ without relying on an EBM, which leads to the  alternating back-propagation training scheme \cite{ABP_aaai}. $G_\alpha(X,h)$ directly learns from observed training data rather than synthesized examples provided by an EBM.   
Compared with $G_\alpha(X,h)$ trained independently, ''Ours\_F`` achieves better performance, which~validates~the effectiveness of the cooperative learning strategy.

\begin{table*}[t!]
  \centering
  \scriptsize
  \renewcommand{\arraystretch}{1.0}
  \renewcommand{\tabcolsep}{0.7mm}
  \caption{Performance comparison with alternative uncertainty estimation methods.
  }
  \begin{tabular}{l|cccc|cccc|cccc|cccc|cccc|cccc}
  \hline
% \toprule
  &\multicolumn{4}{c|}{DUTS}&\multicolumn{4}{c|}{ECSSD}&\multicolumn{4}{c|}{DUT}&\multicolumn{4}{c|}{HKU-IS}&\multicolumn{4}{c|}{PASCAL-S}&\multicolumn{4}{c}{SOD} \\
    Method & $S_{\alpha}\uparrow$&$F_{\beta}\uparrow$&$E_{\xi}\uparrow$&$\mathcal{M}\downarrow$& $S_{\alpha}\uparrow$&$F_{\beta}\uparrow$&$E_{\xi}\uparrow$&$\mathcal{M}\downarrow$& $S_{\alpha}\uparrow$&$F_{\beta}\uparrow$&$E_{\xi}\uparrow$&$\mathcal{M}\downarrow$& $S_{\alpha}\uparrow$&$F_{\beta}\uparrow$&$E_{\xi}\uparrow$&$\mathcal{M}\downarrow$& $S_{\alpha}\uparrow$&$F_{\beta}\uparrow$&$E_{\xi}\uparrow$&$\mathcal{M}\downarrow$& $S_{\alpha}\uparrow$&$F_{\beta}\uparrow$&$E_{\xi}\uparrow$&$\mathcal{M}\downarrow$ \\ \hline   
   CVAE & .890 & .849 & .925 & .036 & .919 & .918 & .948 & .034 & .836 & .761 & .868 & .056 & .918 & .906 & .955 & .028 & .863 & .835 & .902 & .062  & .838 & .830 & .878 & .071\\ 
   CGAN & .888 & .849 & .927 & .035 & .917 & .914 & .944 & .036 & .837 & .764 & .871 & .054 & .917 & .908 & .955 & .028 & .865 & .839 & .906 & .059  & .836 & .833 & .874 & .072\\ 
  MCD & .881 & .842 & .918 & .038 & .917 & .917 & .944 & .036 & .828 & .753 & .859 & .057 & .915 & .908 & .951 & .030 & .863 & .837 & .902 & .062 & .834 & .831 & .868 & .073  \\ 
  ENS & .885 & .841 & .921 & .037 & .921 & .917 & .948 & .035 & .831 & .752 & .862 & .057 & .916 & .901 & .952 & .030 & .858 & .827& .897 & .065 & .835 & .828 & .872 & .073 \\ \hline
  \textbf{Our\_F} & \textbf{.902} & \textbf{.877} & \textbf{.936} & \textbf{.032} & \textbf{.928} & \textbf{.935} & \textbf{.955} & \textbf{.030} & \textbf{.857} & \textbf{.798} & \textbf{.889} & \textbf{.049} & \textbf{.927} & \textbf{.917} & \textbf{.960} & \textbf{.026} & \textbf{.873} & \textbf{.846} & \textbf{.909} & \textbf{.058}  & \textbf{.854} & \textbf{.850} & \textbf{.885} & \textbf{.064}\\\hline
  \end{tabular}
  \label{tab:alternative_sampling_processmodeling__models}
  \vspace{-3mm}
\end{table*}

\begin{figure}[!t]
   \begin{center}
   \begin{tabular}{c@{ }}
   {\includegraphics[width=0.96\linewidth]{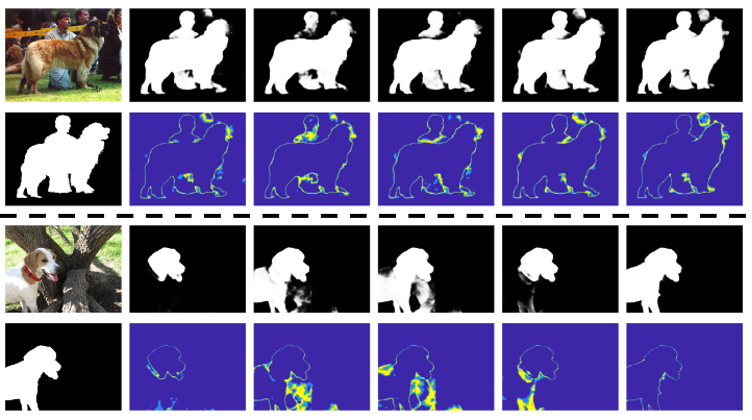}}
        \\
        % (b) \\
   \end{tabular}
   \end{center}
%   \vspace{-2mm}
\caption{\small Saliency predictions of alternative uncertainty estimation methods.
% Images in the 
For each panel, the first row shows an input image followed by the mean predictions of different alternative uncertainty estimation methods and ours, and the second row shows the ground truth saliency map followed by the uncertainty maps of different methods. From left to right columns, they are Image/ground truth, \enquote{CVAE}, \enquote{CGAN}, \enquote{MCD}, \enquote{ENS} and ours.
}
\vspace{-2mm}
   \label{fig:visualization_mean_sigma}
\end{figure}

\textbf{Design of encoder and encoder structures:}
We replace the decoder part in our proposed framework by the one of those existing deterministic saliency prediction methods, \eg, ITSD \cite{itsd_sal}. We select ITSD \cite{itsd_sal} because of the availability of its code and the state-of-the-art performance. We show its performance in Table \ref{tab:ablation_study} as \enquote{ITSD\_Ours}. Further, we replace the ResNet50 \cite{ResHe2015} encoder backbone in our model by VGG16 \cite{VGG} and denote the new model as \enquote{VGG16\_Ours}. The consistently better performance of \enquote{ITSD\_Ours} than the original \enquote{ITSD} validates the superiority of the generative cooperative learning framework. Comparable performances are observed in our models with different backbone selections.

\subsection{Alternative Uncertainty Estimation Methods}
%\subsection{Alternative Generative Frameworks}
\label{alternative_generative_models}
In this section, we compare our generative framework with other alternative uncertainty estimation methods for saliency prediction. We first design two alternatives based on CVAEs \cite{cvae} and CGANs \cite{conditional_gan}, respectively. For the CVAE model, we follow  \citet{ucnet_jornal}, except that we replace its decoder network by our decoder \cite{Ranftl2020}.
As to the CGAN, we optimize
the adversarial loss \cite{GAN_nips} of the conditional generative adversarial network that consists of a conditional generator $G$ and a conditional discriminator $D$. Specifically, we use the same latent variable model $G_\alpha(X,h)$ as that in our model 
% showing in Figure \ref{generator_net} 
for the generator of CGAN.
For the discriminator, we design a fully~convolutional~discriminator as in \citet{Hung_semiseg_2018} to classify each pixel into real (ground truth) or fake (prediction).
To train CGAN, the discriminator $D$ is updated with the discriminator loss $\mathcal{L}_{ce}(D(Y),\mathbf{1})+\mathcal{L}_{ce}(D(G_\alpha(X,h)),\mathbf{0})$, where $\mathcal{L}_{ce}$ is the binary cross-entropy loss, $\mathbf{1}$ and $\mathbf{0}$ are all-one and all-zero maps of the same spatial size as $Y$. $G_{\alpha}$ is updated with $\mathcal{L}_{ce}(G_\alpha(X,h),Y)+\lambda_d \mathcal{L}_{ce}(D(G_\alpha(X,h)),\mathbf{1})$,~where %$\mathcal{L}_s$ is the structure-aware loss in \citet{F3Net_aaai2020}, and 
$\mathcal{L}_{ce}(D(G_\alpha(X,h)),\mathbf{1})$ is the adversarial loss for $G$ and $\mathcal{L}_{ce}(G_\alpha(X,h),Y)$ is the cross-entropy loss between the outputs of $G$ and the observed saliency maps.~We~set~$\lambda_d=0.1$.

We also design two ensemble-based saliency detection frameworks with Monte Carlo dropout \cite{dropout_bayesian} and deep ensemble \cite{deep_ensemble} to produce multiple predictions,
% with a multi-head decoder, 
and show their performance as
% . Performance of these two models is shown as
\enquote{MCD} and \enquote{ENS}
% in \ref{alternative_generative_models} respectively 
in Table \ref{tab:alternative_sampling_processmodeling__models} respectively. For \enquote{MCD}, we add dropout to each level of features of the encoder within the noise-free encoder-decoder $G_\alpha(X)$ with a dropout rate 0.3, and use dropout in both of  training and testing processes. For \enquote{ENS}, we attach five MiDaS decoder \cite{Ranftl2020} to $G_\alpha(X)$, which are initialized differently, leading to five outputs of predictions. For both ensemble-based frameworks, similar to our generative models, we use the mean prediction averaging over 10 samples in testing as the final prediction, and the entropy of the mean prediction as the predictive uncertainty following \citet{detlefsen2019reliable}.

We show performance of alternative uncertainty estimation models in Table \ref{tab:alternative_sampling_processmodeling__models}, and visualize the mean prediction and predictive uncertainty for each method in Figure~\ref{fig:visualization_mean_sigma}. For the CVAE-based framework, designing the approximate inference network takes extra efforts, and the imbalanced inference model may lead to the posterior collapse issue as discussed in \citet{Lagging_Inference_Networks}. For the CGAN-based model, according to our experiments, the training is sensitive to the proportion of the adversarial loss. Further, it cannot infer the latent variables $h$, which makes the model hard to learn from incomplete data for weakly supervised learning. For the deep ensemble \cite{deep_ensemble} and MC dropout \cite{dropout_bayesian} solutions, they can hardly improve model performance, although the produced predictive uncertainty maps can explain model prediction to some extent.
Compared all above alternative methods, our proposed framework is stable due to maximum likelihood learning, and we can infer latent variables $h$ without the need of an extra encoder. Further, as we directly sample from the truth posterior distribution via Langevin dynamics, instead of the approximated inference network, we have more reliable and accurate predictive uncertainty maps compared with other alternative solutions.

\section{Conclusion and Discussion}
In this paper, we propose a novel energy-based generative saliency prediction framework based on the conditional generative cooperative network, where a conditional latent variable model and an conditional energy-based model are jointly trained in a cooperative learning scheme to achieve a coarse-to-fine saliency prediction. The latent variable model serves as a coarse saliency predictor that provides a fast initial saliency prediction, while the energy-based model serves as a fine saliency predictor that further refines the initial output by the Langevin revision. Even though each of the models can represent the conditional probability distribution of saliency, the cooperative representation and training can offer the best of both worlds. Moreover, we propose a \textit{cooperative learning while recovering} strategy and apply the model to the weakly supervised saliency detection scenario, in which partial annotations (\eg, scribble annotations) are provided for training. As to the cooperative recovery part of the proposed strategy, the latent variable model serves as a fast but coarse saliency recoverer that provides an initial recovery of the missing annotations from the latent space via inference process, while the energy-based model serves as a slow but fine saliency recoverer that refines the initial recovery results by Langevin dynamics. Combining these two types of recovery schemes leads to a coarse-to-fine recoverer. Further, we find that the learned energy function
in the energy-based model can serve as a refinement module, which can be easily plugged into the existing pre-trained saliency prediction models. The energy function is the potential cost function trained from the saliency prediction task. In comparison to a mapping function from image to saliency, the cost function captures the criterion to measure the quality of the saliency given an image, and is more generalizable so that it can be used to refine other saliency predictions. Extensive results exhibit that, compared with both conventional deterministic mapping methods and alternative uncertainty estimation methods, our framework can lead to both accurate saliency predictions for computer vision tasks and reliable uncertainty maps indicating the model confidence in performing saliency prediction from an image. As to a broader impact, the proposed computational framework might also benefit the researchers in the field of computational neuroscience who investigate human attentional mechanisms. The proposed coarse-to-fine saliency prediction model and recovery model may shed light on a clear path toward the understanding of relationship between visual signals and human saliency.

\bibliography{mybibfile}

\end{document}